\documentclass[twoside,11pt]{article}

\usepackage{jmlr2e}
\usepackage{amsmath}
\usepackage{amssymb}
\usepackage{amsfonts}
\usepackage{bbm}
\usepackage{graphicx}
\usepackage[ruled,vlined, linesnumbered]{algorithm2e}

\jmlrheading{0}{0000}{0-0}{0/00}{0/00}{xxxx}{Benjamin Letham and Eytan Bakshy}

\ShortHeadings{Bayesian Optimization for Policy Search via Online-Offline Experimentation}{Letham and Bakshy}
\firstpageno{1}

\begin{document}

\title{Bayesian Optimization for Policy Search via Online-Offline Experimentation}

\author{\name Benjamin Letham \email bletham@fb.com
\AND
\name Eytan Bakshy \email eytan@fb.com \\
\addr Facebook\\Menlo Park, California, USA
}

\editor{}

\maketitle

\begin{abstract}
Online field experiments are the gold-standard way of evaluating changes to real-world interactive machine learning systems.  Yet our ability to explore complex, multi-dimensional policy spaces---such as those found in recommendation and ranking problems---is often constrained by the limited number of experiments that can be run simultaneously. To alleviate these constraints, we augment online experiments with an offline simulator and apply multi-task Bayesian optimization to tune live machine learning systems. We describe practical issues that arise in these types of applications, including biases that arise from using a  simulator and assumptions for the multi-task kernel. We measure empirical learning curves which show substantial gains from including data from biased offline experiments, and show how these learning curves are consistent with theoretical results for multi-task Gaussian process generalization. We find that improved kernel inference is a significant driver of multi-task generalization. Finally, we show several examples of Bayesian optimization efficiently tuning a live machine learning system by combining offline and online experiments.
\end{abstract}


\section{Introduction}\label{sec:intro}

Online field experiments, commonly referred to as ``A/B tests,'' are the gold-standard method for evaluating changes to Internet services and are an important tool for driving improvements \citep{kohavi2007practical,manzi2012uncontrolled,bakshy2014}.  Such experiments typically test only a few possible discrete variants, such as the presence or absence of a new feature, or a test of a new ranking model. Bayesian optimization \citep{jones98} is an efficient approach to exploring and optimizing large, continuous parameter spaces in noisy environments, including field experiments~\citep{letham2017constrained}. These methods open the possibility of solving reinforcement learning-type problems with Bayesian optimization~\citep{lizotte2007gps4rl, wilson2014bo4rl, calandra16, metzen16, marco17} in the context of real-world machine learning systems.

Bayesian optimization works by fitting a response surface model to a set of evaluated design points (e.g., a parameter configuration that determines the behavior of software, such as a policy) and iteratively deploying new points based on an explore/exploit algorithm. A practical challenge in tuning online machine learning systems with Bayesian optimization is that each online field experiment can require millions of samples due to low signal-to-noise ratios in target outcomes \citep{kohavi14}. Given a limited available population, it can then be challenging to run enough experiments and evaluate enough design points to yield an accurate response surface model and to effectively optimize many parameters simultaneously.

Offline simulators can provide a higher-throughput mechanism for evaluating the effects of changes to production systems. For Internet services, an offline simulator can provide estimates of humans' interactions with a machine learning system by replaying sessions to the system under an alternative configuration policy. Behavioral models are then used to simulate events and compute aggregate outcomes (e.g., average cumulative rewards) to provide a counterfactual estimate of the alternative policy~\citep{bottou2013counterfactual,dudik2014doubly}.

There are many sources of bias in simulators that can limit their utility.  Even in cases where accurate predictive models are available, as the policy under evaluation pushes system behavior further from the status quo, it can be increasingly difficult to model individuals' responses to off-policy states that may be very far from previously observed states. Additionally, unobserved confounding can make it difficult to obtain high-quality counterfactual estimates. For example, \citet{bottou2013counterfactual} described how estimation of many interventions of interest in a production advertising system at Microsoft required careful consideration of the causal structure of the underlying system.  Such analyses are often difficult to perform adequately in arbitrary machine learning systems that may not have randomized assignment policies.

It generally will not be possible to accurately simulate the interactions between humans and a machine learning system, especially for outcomes with more complex dynamics than click-through-rates.  Take for example the case of ranking stories shown to people in the Facebook News Feed.  The number of posts an individual is exposed to, and which posts an individual interacts with within the span of a single session, will be influenced by the composition (e.g., videos, images, text) and ordering of posts.  Any interactions with specific content may in turn affect patterns of activity in present and future sessions~\citep{eckles2016estimating}.
Techniques from reinforcement learning and causal inference can be used to improve simulator accuracy in specific applications, including some multi-armed bandit settings \citep{tang13, hofmann13, li15b, li15a}, but generally these dynamics will not be captured.
As a result, offline experiments are often considered as a precursor to evaluating a machine learning system with online experiments \citep{shani2011}. Here we show how they can be used in tandem with online experiments to increase the accuracy and throughput of online Bayesian optimization.

Bayesian optimization with the aid of a simulator is a special case of multi-fidelity optimization, in which there is a target function to be optimized along with a collection of cheaper approximations of varying fidelity. By modeling the relationship between the approximations and the target function, the approximations can be used to more efficiently explore the design space and to accelerate the optimization.  \citet{swersky13} developed multi-task Bayesian optimization (MTBO), in which a multi-task Gaussian process (MTGP) is used to model a response surface jointly across tasks, which could be a discrete set of fidelity levels. They showed that MTBO is effective for multi-fidelity hyperparameter optimization with downsampled datasets providing low-fidelity estimates.

In this paper, we provide a rigorous, empirical evaluation of how MTBO can be used effectively for real-world policy search and configuration problems.  We first describe the empirical context of this work (Section \ref{sec:context}), which is an online ranking system in a non-stationary, noisy environment (i.e., live Facebook traffic). We describe a simple simulator of that environment which uses the existing event prediction models to provide counterfactual estimates. In Section \ref{sec:gp} we describe the MTGP model that enables borrowing of strength from offline measurements to online measurements, and review the assumptions behind the ICM kernel that we use. Section \ref{sec:mtgp_emp} provides the first set of empirical results on the suitability of the MTGP for response surface modeling in an online ranking system. Importantly, we show that with the MTGP we can directly use a simple, biased simulator together with a small number of online experiments to accurately predict online outcomes. In Section \ref{sec:optimization} we evaluate MTGPs and MTBO on a corpus of real-world experiments conducted using our proposed methodology. These results show that the addition of the simulator leads to better user experiences during the optimization and an improved final state. Finally, in Section \ref{sec:generalization} we review theoretical results on MTGP generalization, and then construct empirical learning curves for generalization in this setting. We provide an empirical analysis of the factors behind MTGP generalization and show how theory provides practical guidance on best practices for policy search with combined online and offline experimentation.

\section{Empirical Context and the Simulator}\label{sec:context}
In this paper we describe the tuning of a particular recommendation system, the Facebook News Feed. Generally, recommendation systems consist of a number of machine learning models that predict different events, like commenting or clicking. When an item is sent to the system for scoring, the score is based on these event predictions as well as features of the item, such as the strength of connection between viewer and content producer. The actual score is computed by applying a collection of rules and weights, which encode business logic and provide a way of combining all of the event predictions into an overall score. This collection of rules is called the \textit{value model}, and can be seen as parameterizing the ranking policy used in a recommender system. Individual terms in the value model may target statistical surrogates~\citep{prentice1989surrogate,athey2016estimating} that are expected to improve long-term outcomes of interest that are not directly modeled by event-based models that measure more immediate rewards. The value model forms a convenient way of balancing multi-objective decision problems faced in industrial settings, and similar approaches can be found at Yahoo!, LinkedIn, and other organizations \citep{agarwal2011click,agarwal2015constrained}.

A typical value model contains many parameters that affect its performance, such as weights for combining event prediction probabilities and rules that apply additional weights to the scores under certain conditions. These rules and weights constitute the policy for the system. The policy directly impacts long-term individual-level outcomes and must be tuned to achieve the desired system performance. The outcomes of interest can only be measured with online field experiments, and so the policy search must be done online.  For example, in the case of News Feed ranking, a team may wish to tune value model parameters to maximize a measure of meaningful interactions \citep{fbnews}, subject to constraints on other engagement and ecosystem-related outcomes. This is the type of policy search problem that we wish to solve using Bayesian optimization.

We use the term \textit{experiment} to refer to the process of tuning a particular set of parameters. An experiment represents a policy search problem and comprises potentially many A/B tests and simulator runs. The set of parameters being tuned and their ranges form the design space for the experiment, and we refer to the parameter values of any particular configuration as a \textit{policy}. When policies are evaluated, either online or offline, we evaluate several in parallel and refer to the policies being simultaneously evaluated as a \textit{batch}. For each evaluation, we record measurements of several \textit{outcomes}, which are summary statistics of the various measurements of system performance (including individuals' behaviors). These measured outcomes of either an online or offline test are referred to as \textit{observations} for those policies.

Because of the limitations in online experimentation throughput, policy search would be greatly accelerated if the outcomes of interest could be measured offline, via simulation. In fact, we already have predictive models for the event probabilities of many outcomes of interest, which are incorporated in the value model. A na\"{\i}ve approach to offline counterfactual estimation is to replay user sessions to the recommender system under a modified value model, apply the event models to the items selected by the system (e.g., probability of comment), and aggregate the model predictions across many replayed sessions to estimate top-level outcomes (e.g., total comments).

\begin{figure}[tb]
\centering
\includegraphics{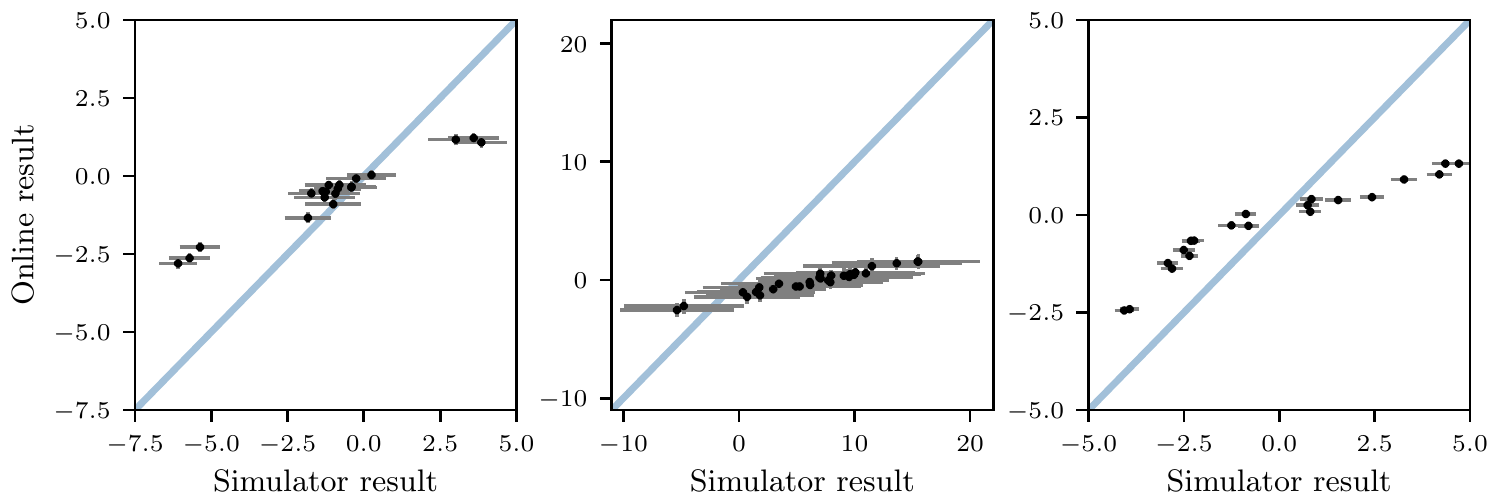}
\caption{Simulator observations compared to results of online tests for three outcomes (panels). Each marker shows a policy that was evaluated in both the simulator and with an online test. The simulator measurements were correlated with the actual online results, but were too biased to be used in the place of online experiments.
}
\label{fig:online_offline}
\end{figure}

This simulation approach is attractive because it requires little additional infrastructure by making use of existing event prediction models, and it can evaluate arbitrary changes to the policy. However, as described in the Introduction, it suffers from significant limitations that prevent it from producing accurate estimates. Examples of the bias between offline estimates with this technique and online results are shown in Fig. \ref{fig:online_offline}. This figure shows three outcomes from experiments that varied value model parameters. In each experiment, a batch of 20 policies were evaluated both in the simulator and with online tests. Fig. \ref{fig:online_offline} compares the observations of each policy, online and with the simulator.

Observations in Fig. \ref{fig:online_offline} are shown relative to the status quo, meaning $0$ on each axis corresponds to the status quo value. Generally the simulator observations were correlated with the observations of online experiments, but showed increasingly large discrepancies as the outputs deviated further from the status quo. Some of this is off-policy bias to be expected with this simulation approach since it uses only on-policy trained predictive models. There are also relevant dynamics of the system that are not being simulated that provide another source of bias. Since some of these outcomes will be used as constraints in the policy search problem, the monotonic relationship between simulator and online results is not sufficient to do the optimization with the simulator alone. Moreover, the bias between simulator and online results depends on the experiment (which parameters are being tuned), so any adjustment must be learned separately for each experiment.  In general, we observe that simulators tend to systematically overestimate the size of effects, likely because of the difficulties with modeling elasticities in time and attention \citep{simon1971designing,simon1996sciences}. We expect to see similar behaviors in other settings, such as markets, that may be subject to difficult-to-model elasticities.

The methodology that we develop here allows for these offline estimates to be directly combined with online observations to improve the throughput of online policy search, despite the bias. We now describe the MTGP model for online-offline optimization. While the empirical setting of this work is value model tuning, many other parameters in information ranking and retrieval systems induce changes in system behavior that can be estimated offline in a similar way. The methodology can be applied to many such other policy search problems.

\section{Response Surface Modeling with the Gaussian Process}\label{sec:gp}

Bayesian optimization relies on modeling the outcome of an online test as a function of policy parameters. A typical approach is to use a Gaussian process (GP) prior as the response surface model. The GP prior, 
$f \sim \mathcal{GP}(m(\cdot), k(\cdot, \cdot))$, 
is defined by a mean function $m(\mathbf{x}) = \mathbb{E}[f(\mathbf{x})]$ and a covariance function $k(\mathbf{x}, \mathbf{x}') = \textrm{Cov}[f(\mathbf{x}), f(\mathbf{x}')]$. The covariance function specifies the covariance between any two points in the design space and is typically chosen so that the covariance decays with distance, such as in the ARD RBF kernel
\begin{equation*}
    k(\mathbf{x}, \mathbf{x}') = \tau^2 \exp\left( -\frac{1}{2} \sum_{j=1}^m \left(\frac{x_j - x'_j}{\ell_j} \right)^2 \right).
\end{equation*}
The kernel induces smoothness in $f(\mathbf{x})$ since points that are nearby in space are given high covariance. The degree of smoothness depends on the kernel variance and lengthscales, $\tau^2$ and $\ell_j$, which are inferred from data, typically by maximizing the marginal likelihood or with slice sampling \citep{murray10}.

Under a GP, any finite collection of function evaluations $f(\mathbf{x}_1), \ldots, f(\mathbf{x}_n)$ are jointly normally distributed. For notational convenience, let $X = [\mathbf{x}_1, \ldots, \mathbf{x}_n]$ denote the evaluated points,  $m(X) = [m(\mathbf{x}_1), \ldots, m(\mathbf{x}_n)]^\top$ denote the vector of means, and
\begin{equation*}
    K(X, X) =
    \begin{pmatrix}
    k(\mathbf{x}_1, \mathbf{x}_1) & \cdots & k(\mathbf{x}_1, \mathbf{x}_n) \\
    \vdots & \ddots & \vdots \\
    k(\mathbf{x}_n, \mathbf{x}_1) & \cdots & k(\mathbf{x}_n, \mathbf{x}_n)
    \end{pmatrix}
\end{equation*}
denote the covariance matrix across $X$. Then, the GP prior has
\begin{equation*}
f(\mathbf{x}_1), \ldots, f(\mathbf{x}_n) \sim \mathcal{N}(m(X), K(X, X)).
\end{equation*}
We typically do not directly observe the ``true" outcome $f(\mathbf{x}_i)$, rather we observe a noisy measurement $y_i = f(\mathbf{x}_i) + \epsilon_i$, where $\epsilon_i \sim \mathcal{N}(0, \eta_i^2)$. For online field experiments, we measure an aggregate outcome over a large test population and so are able to directly measure both the mean outcome $y_i$ and its sample variance $\eta_i^2$. Let $\mathbf{y} = [y_1, \ldots, y_n]$ and $\boldsymbol{\eta}^2 = [\eta_1^2, \ldots, \eta_n^2]$. It follows that
\begin{equation*}
    \mathbf{y} \sim \mathcal{N}(m(X), K(X, X) + \textrm{diag}(\boldsymbol{\eta}^2)).
\end{equation*}
Given observations $\{X, \mathbf{y}\}$, Gaussian closure under conditioning leads to an analytic posterior for the function value at any $\mathbf{x}_*$
\begin{equation}\label{eq:gp_posterior}
    f(\mathbf{x}_*) |\mathbf{x}_*, X, \mathbf{y} \sim \mathcal{N}(\mu(\mathbf{x}_*, X, \mathbf{y}), \sigma^2(\mathbf{x}_*, X)),
\end{equation}
where
\begin{subequations}
\begin{align}\label{eq:kriging_a}
    \mu(\mathbf{x}_*, X, \mathbf{y})  & = m(\mathbf{x}_*) + K(\mathbf{x}_*, X)[K(X, X) + \textrm{diag}(\boldsymbol{\eta}^2)]^{-1} (\mathbf{y} - m(X)), \\\label{eq:kriging_b}
    \sigma^2(\mathbf{x}_*, X) &= k(\mathbf{x}_*, \mathbf{x}_*) - K(\mathbf{x}_*, X)[K(X, X) + \textrm{diag}(\boldsymbol{\eta}^2)]^{-1} K(X, \mathbf{x}_*),
\end{align}
\end{subequations}
with $K(\mathbf{x}_*, X)$ the $1 \times n$ vector of covariances between $\mathbf{x}_*$ and each point in $X$.

An interesting property of GPs that can be seen from (\ref{eq:kriging_b}) is that the posterior variance does not depend on $\mathbf{y}$, a fact that will be useful for characterizing generalization. GP regression was largely developed in the geostatistics community, where (\ref{eq:kriging_a}) and (\ref{eq:kriging_b}) are referred to as the kriging equations \citep{cressie90}. \citet{rasmussen06} provide a thorough machine learning perspective on GP regression.

\begin{figure}[tb]
\centering
\includegraphics{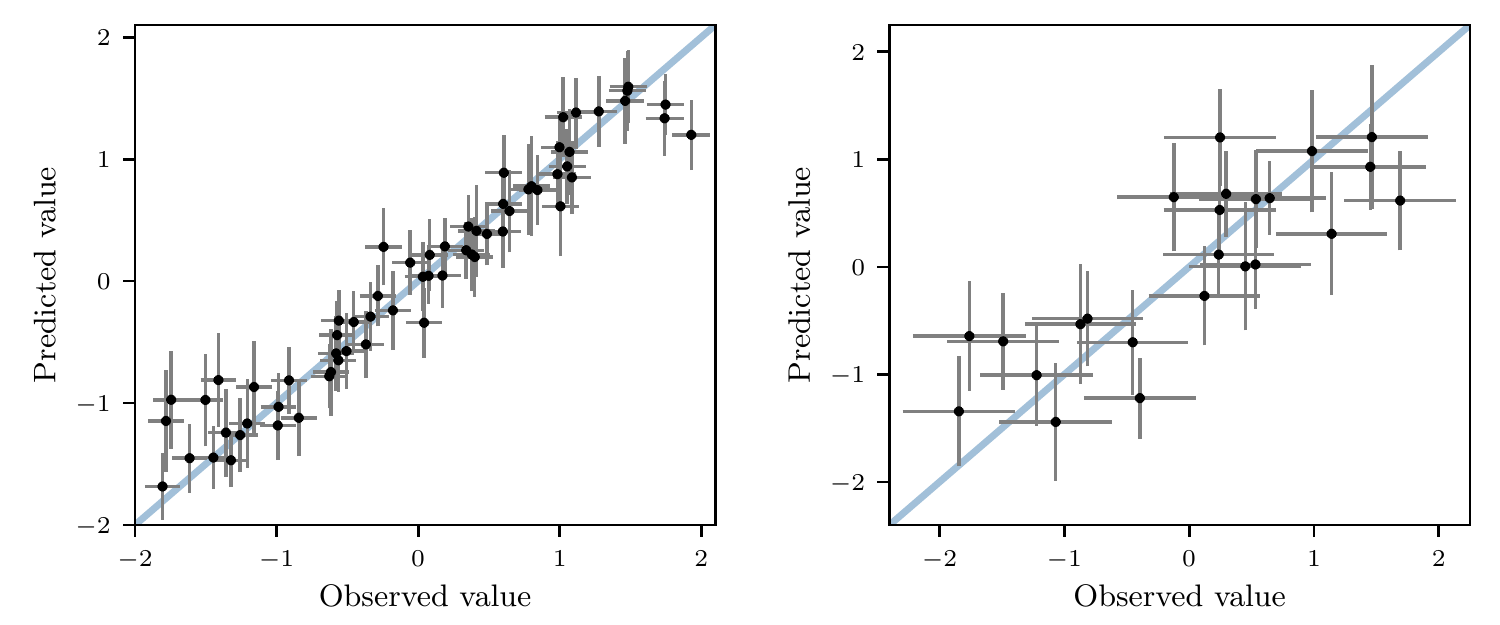}
\caption{Leave-one-out cross validation predicting the outcome of online tests with a GP. (Left) A GP fit to 64 observations in a 9-dimensional value model tuning experiment. (Right) A GP fit to 24 observations in a 3-dimensional video playback controller policy tuning experiment. GPs provide good predictions for online field experiment outcomes, making them a suitable model for Bayesian optimization.}
\label{fig:gp_loo}
\end{figure}

The GP has several properties that make it especially useful as a response surface model for Bayesian optimization. First, it provides uncertainty estimates for the function value at unobserved points. This is critical for applying an explore/exploit algorithm since exploration is driven by the uncertainty estimates. Second, both mean and variance predictions are available in closed form, in (\ref{eq:kriging_a}) and (\ref{eq:kriging_b}). This allows for fast gradient optimization of functions of the posterior when trying to identify the optimal point to test next. Finally, the smoothness assumption implied by the kernel tends in practice to be a good fit for the response surfaces of systems parameter tuning experiments. We have used the GP for parameter tuning via A/B tests in a variety of settings and consistently found good predictive performance \citep{letham2017constrained}. Fig. \ref{fig:gp_loo} shows examples of using a GP to predict online outcomes for two online policy search problems: a value model tuning experiment, and optimizing parameters of a video playback controller policy. For each of these, a batch of policies were tested online, and a GP was fit to the observations. For these experiments, leave-one-out cross-validation shows that the GP was able to make accurate out-of-sample predictions of the results of online tests. This model performance is typical of many successful Bayesian optimization experiments that we have run.

\subsection{The Multi-Task Gaussian Process}\label{sec:mtgp}

Bayesian optimization with a GP response surface model is effective for policy search in online systems, however the number of observations (online tests) required for good GP predictions with 10--24 parameters can be prohibitive for systems with limited capacity for online experiments.  The MTGP provides a natural way to include simulator observations into the model and reduce the number of online observations needed.

The MTGP extends the GP from a single function $f$ to a collection of functions $f_1, \ldots, f_D$, each of these the response surface for a task. For our offline-online policy search, we initially have two tasks: an online task consisting of the online observations, and an offline task with the simulator observations. In Section \ref{sec:optimization} we will also model different simulator batches with separate tasks as well, and so will describe the model for an arbitrary number of tasks $D$.

The MTGP covariance function models the covariance across tasks in addition to the covariance across the design space: $k((d, \mathbf{x}), (d', \mathbf{x}')) = \textrm{Cov}[f_d(\mathbf{x}), f_{d'}(\mathbf{x}')]$. We will make two assumptions to construct a multi-task covariance function. The first assumption is that all tasks share the same spatial kernel, which we will denote $\kappa(\mathbf{x}, \mathbf{x}')$. This assumption is especially useful because it allows us to rely on the simulator to estimate the spatial kernel hyperparameters, which can have many more observations than the online task. A similar approach is taken in reinforcement learning in robotics, where information sharing across tasks is facilitated by sharing a kernel \citep{chai08}. The second assumption is separability between the task covariance and the spatial covariance, which assumption yields
\begin{equation}\label{eq:icm_cov}
    \textrm{Cov}[f_d(\mathbf{x}), f_{d'}(\mathbf{x}')] = B_{d, d'} \kappa(\mathbf{x}, \mathbf{x}').
\end{equation}
Here $B$ is a positive semidefinite task covariance matrix with element $B_{d, d'}$ the covariance between tasks $d$ and $d'$, and $\kappa(\mathbf{x}, \mathbf{x}')$ is the parameter covariance function. This is called the intrinsic coregionalization model (ICM) \citep{alvarez11}. The covariance function (\ref{eq:icm_cov}) can be substituted into (\ref{eq:kriging_a}) and (\ref{eq:kriging_b}) to immediately extend GP regression to the multi-task setting. The MTGP maintains all of the desirable properties of the GP for Bayesian optimization, while also modeling the relationship between the online and offline tasks.

The ICM kernel implicitly models each task function, $f_d$, as being a linear combination of independent latent functions. In the case of two tasks, it models each task as a sum of two latent functions
\begin{subequations}
\begin{align}\label{eq:icm_latent_a}
    f_1(\mathbf{x}) &= a_{11} u_1(\mathbf{x}) + a_{12} u_2(\mathbf{x}), \\\label{eq:icm_latent_b}
    f_2(\mathbf{x}) &= a_{21} u_1(\mathbf{x}) + a_{22} u_2(\mathbf{x}),
\end{align}
\end{subequations}
where $u_1(\mathbf{x})$ and $u_2(\mathbf{x})$ have GP priors with the same covariance function $\kappa(\mathbf{x}, \mathbf{x}')$ \citep{alvarez12}. The weights are related to the cross-task covariance of (\ref{eq:icm_cov}) as $B_{12} = a_{11} a_{21} + a_{12} a_{22}$. High cross-task covariance thus corresponds to the two task functions concentrating their weight on the same latent function. In the extreme case of $a_{12} = a_{21} = 0$, the ICM kernel corresponds exactly to modeling the two tasks with independent GPs that have the same covariance function.

This formulation gives insight into what type of simulator biases the MTGP with ICM kernel will be capable of capturing. Rearranging (\ref{eq:icm_latent_a}) and (\ref{eq:icm_latent_b}), we have that $f_2(\mathbf{x}) = f_1(\mathbf{x}) + u'(\mathbf{x})$ where $u'(\mathbf{x}) = (a_{21} - a_{11}) u_1(\mathbf{x}) + (a_{22} - a_{12}) u_2(\mathbf{x})$. Here $u'(\mathbf{x})$ is the bias from one task to the other, and we see that $u'(\mathbf{x})$ is a GP with covariance function $\kappa(\mathbf{x}, \mathbf{x}')$. Thus, the ICM kernel implicitly models the simulator bias with a GP, and can capture any simulator bias that has the same degree of smoothness as the response surface itself. The biases shown in Fig. \ref{fig:online_offline} are smooth over the output and suggest that an ICM kernel should be successful at modeling and thus adjusting for the bias seen with this simulator. 

MTGP inference requires fitting the hyperparameters of the spatial covariance function, as in regular GP regression, along with the cross-task covariance matrix $B$. This is done by reparameterizing the cross-task covariance matrix with its Cholesky decomposition $B = LL^\top$, to ensure positive semidefiniteness. The matrix $L$ is then chosen to maximize marginal likelihood. We inferred all kernel hyperparameters ($B$ and the spatial kernel hyperparameters $\tau^2$ and $\ell_j$) from the data by maximizing marginal likelihood, using the Scipy interface to L-BFGS \citep{byrd1995limited, zhu1997algorithm}.

Estimating $B$ requires fitting $D(D+1)/2$ parameters. In a Bayesian optimization setting with relatively few function evaluations, for even moderate $D$ there can be more parameters than can reliably be fit. \citet{bonilla07} proposed using an incomplete Cholesky decomposition
\begin{equation*}
B = \tilde{L}\tilde{L}^\top,
\end{equation*}
where $\tilde{L}$ is a $D \times P$ matrix, $P < D$. This reduces the number of parameters from $\mathcal{O}(D^2)$ to $\mathcal{O}(DP)$ and induces the constraint $\textrm{rank}(B) \leq P$. They show that in practice lower rank parameterizations can have lower generalization error than higher rank parameterizations. For two tasks (online and offline) this is not necessary, but later in this paper we will model different batches with a task and will make use of a low-rank kernel. The rank of the task covariance matrix corresponds to the number of latent functions in (\ref{eq:icm_latent_a}) and (\ref{eq:icm_latent_b}). Those equations correspond to a rank-2 task covariance matrix while a rank-1 task covariance matrix would imply each task function is a scaled version of the same, single latent function.

There are many alternatives to the ICM kernel with more complex structures---\citet{alvarez12} provides a review of multi-task kernels. In our application, the ICM assumption of shared spatial covariance is important for enabling inference of the kernel lengthscales. There are not enough online observations to infer a separate ARD kernel for the online task, and so we rely on the shared kernel and the leverage we can gain from simulator observations for hyperparameter inference. In Section \ref{sec:mtgp_emp} we show empirically that the transfer of information via the shared spatial kernel is appropriate in this setting and performs better than restricting the fitting to the online observations only. In Section \ref{sec:generalization} we use empirical learning curves to show that the shared spatial kernel significantly improves generalization.

\section{Multi-Task Response Surface Models with the Simulator}\label{sec:mtgp_emp}

We now study empirically how the MTGP performs for predicting the results of online value model tuning tests, with the benefit of the na\"{\i}ve simulator described in Section \ref{sec:context}. Our study is based on 11 experiments that were run to improve the performance of the particular recommendation system, according to various online metrics. Each experiment optimized different sets of policy parameters, with the dimensionality ranging from 10 to 20. Each optimization began with a batch of 20 online observations and 100 simulator observations. The policies in these batches were generated from separate scrambled Sobol sequences~\citep{owen1998scrambling} using a different seed. For each experiment, nine outcomes were measured which were treated independently for a total of 99 outcomes to be modeled. The data for each outcome were standardized by subtracting the mean of the observed values and dividing by their standard deviation. A 2-task MTGP was fit separately to each of the 99 outcomes, as described in Section \ref{sec:mtgp}.

\begin{figure}[tb]
\centering
\includegraphics{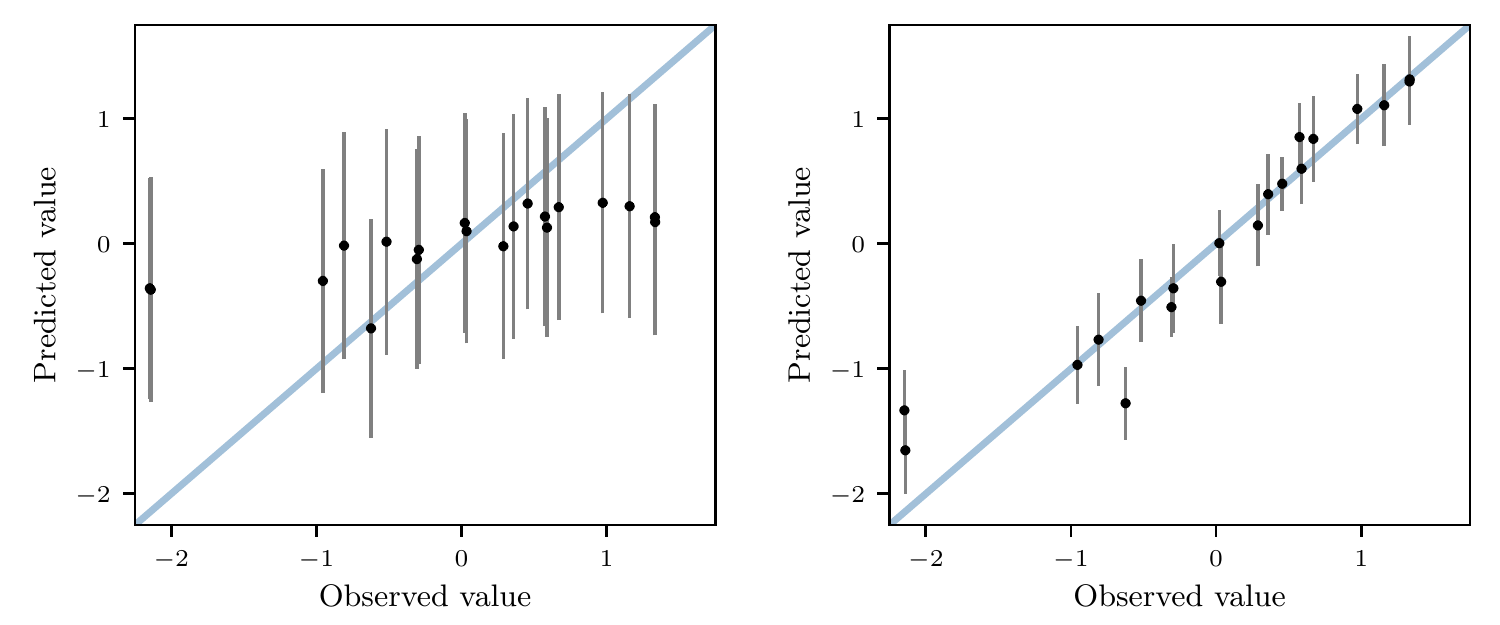}
\caption{Leave-one-out cross validation results predicting the outcome of online tests of 20 policies in a 10-dimensional parameter space. (Left) Results using a GP fit only to the online observations. (Right) Results using an MTGP fit to both the online observations and 100 simulator observations. Due to the relatively high dimensionality, the online observations alone cannot be modeled well enough to make useful predictions, but incorporating the simulator observations allows for accurate out-of-sample predictions.}
\label{fig:loo}
\end{figure}

Across this large body of experiments and models, we have found that including simulator experiments with the MTGP provides significant gains in model performance, for nearly every outcome and experiment. These gains are often substantial enough that the optimization would not be possible without the use of the simulator. Fig. \ref{fig:loo} shows the benefit of the MTGP for one of the outcomes by comparing a single-task GP fit only to the 20 online observations with the MTGP fit to the same 20 online observations plus 100 simulator observations. We evaluated prediction performance using leave-one-out cross-validation, in which each online observation was removed from the data and the model fit to the remaining 19 observations (plus 100 simulator observations for the MTGP) was used to predict the online outcome of the held-out policy. Because the policies deployed online were generating with a space-filling design, this procedure provides a useful evaluation of out-of-sample prediction ability. For these experiments, none of the online policies were included in the simulator batch, so the results for the MTGP are also representative of out-of-sample prediction performance.

The particular experiment shown in Fig. \ref{fig:loo} was over a 10-dimensional parameter space. The cross-validation results for the single-task model show that the 20 online observations alone were not sufficient to be able to make useful predictions in this space. This particular outcome was used as a constraint in the optimization, and the predictive uncertainty is too high to determine if a policy is likely to be feasible or not. Our practical experience with model fits like that of the single-task model is that the policy search would not be successful, and we typically would not continue the experiment. After incorporating simulator experiments with the MTGP, the model performs remarkably well and is able to make accurate predictions.

\begin{figure}[tb]
\centering
\includegraphics{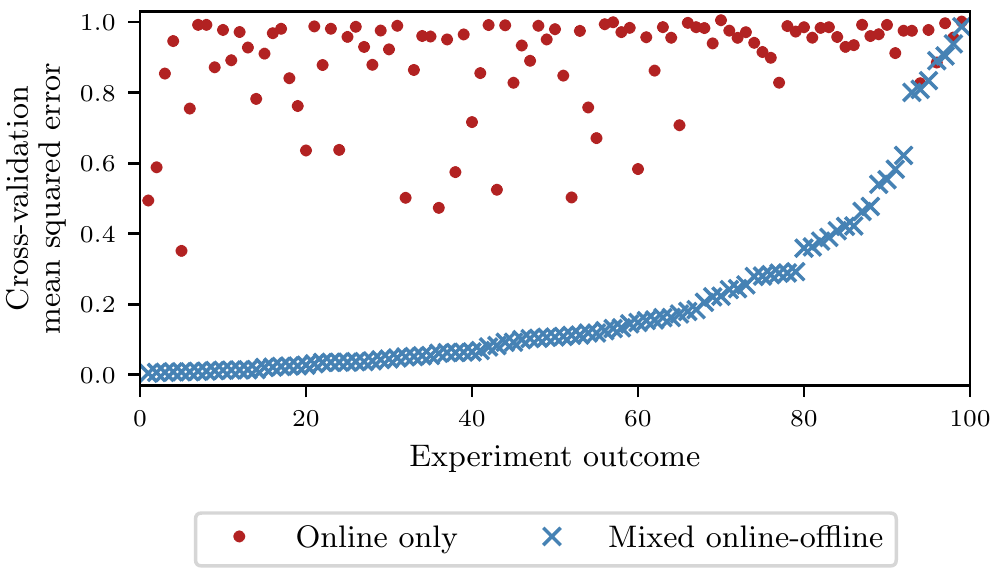}
\caption{For each of 99 experiment outcomes, 20 policies were tested online and 100 policies were tested on the simulator, and two models were separately fit to the observations: a single-task GP to online observations only, and an MTGP to both online and offline observations. Markers show the cross-validation MSE for each of these 99 experiment outcomes, ordered by MTGP MSE. Across this large suite of experiments and outcomes, incorporating simulator observations significantly improved model predictions.}
\label{fig:loo_meta}
\end{figure}

The improvement in model predictions from incorporating simulator observations seen in Fig. \ref{fig:loo} have been seen across many outcomes and experiments. Fig. \ref{fig:loo_meta} shows the mean squared error (MSE) of the 20 online leave-one-out predictions for each of the 99 outcomes described above. Because the data for each outcome were transformed to be mean 0 and standard deviation 1, the expected squared error of a constant prediction at the mean is 1. Fig. \ref{fig:loo_meta} shows that for most of the outcomes, the single-task model with online experiments only did not do much better than predicting the mean. The poor predictions of the single-task GP are not surprising considering there are only 20 observations in 10--20 dimensions. The MTGP, however, produced useful predictions for most outcomes, and importantly, when the prediction errors were high they were still at least as good as the online-only model on all outcomes.

In situations where the simulator performs poorly, as discussed in Section \ref{sec:mtgp} the MTGP contains the single-task GP as a special case and so should in general do no worse than not using the simulator at all, so long as overfitting is avoided. This was the case for all of the outcomes in Fig. \ref{fig:loo_meta}. For those in which the MTGP had high MSE, the lack of substantial improvement over the single-task GP was driven by low correlation between the online and offline tasks. In some instances the low correlation came from a high level of observation noise, while for others it was poor fidelity of the simulator---these factors are analyzed in Online Appendix 1.

\section{Bayesian Optimization by Combining Online and Offline Experiments}\label{sec:optimization}

The ultimate purpose of fitting the MTGP is to produce an accurate model with which to do Bayesian optimization and find an optimal policy. Bayesian optimization with a mix of online and simulator observations proceeds in a similar way as regular single-task Bayesian optimization. We begin with a small number of randomly selected policies to test online, and a much larger set to test offline. The MTGP is fit to the outcomes of these first two batches, as in Section \ref{sec:mtgp_emp}, and is then used to maximize an acquisition function that identifies valuable policies to test in the next batch. We now describe the acquisition function used and how a mix of online and offline batches are used throughout the optimization, and then provide examples of actual value model tuning optimizations done using the method.

\subsection{The Acquisition Function}
To select policies to evaluate with future tests, we use expected improvement integrated over observation noise \citep{letham2017constrained}. The optimization problem that we consider has objective $\max_{\mathbf{x}} f(\mathbf{x})$ and constraints $c_j(\mathbf{x}) \geq 0$, $j=1, \ldots, J$. Here $f$ and $c_j$ represent online outcomes that are modeled with separate MTGPs. Consider first a single-task problem with noisy observations of objective and constraint outcomes at a set of policy parameters $X$. 

We let $\mathcal{D}$ denote the set of all observed data, and use the shorthand $f(X)$ to denote the vector of (unobserved, true) objective function evaluations at each policy in $X$. The unobserved, true constraint evaluations are likewise denoted $\mathbf{c}(X) = [c_1(X), \ldots, c_J(X)]$. Without observation noise, the improvement at $\mathbf{x}$ over the current best feasible policy can be expressed as\footnote{For ease of exposition, the definition given here assumes the existence of a current feasible point. The case where all observations are infeasible is handled in \citet{letham2017constrained}.}
\begin{equation*}
     I(\mathbf{x}, X) = \max \left (0, f(\mathbf{x}) - \max_{\mathbf{x}^* \in X: \mathbf{c}(\mathbf{x}^*) \geq \mathbf{0}} f(\mathbf{x}^*) \right) \mathbbm{1}_{\mathbf{c}(\mathbf{x}) \geq \mathbf{0}}.
\end{equation*}
EI is the expectation over the posterior of $f(\mathbf{x})$ of this quantity, and the noisy EI acquisition function is defined by further integrating over the observation noise,
\begin{equation*}
    \alpha_{\textrm{NEI}}(\mathbf{x} | X, \mathcal{D}) = \mathbb{E}_{f(X), \mathbf{c}(X)} \left[ \mathbb{E}_{f(\mathbf{x})}\left[I(\mathbf{x}, X)  | f(X), \mathbf{c}(X) \right] | \mathcal{D} \right].
\end{equation*}
The inner expectation has an analytic solution. The outer expectation is over the multivariate normal GP posterior and can be effectively handled with numerical (quasi-Monte Carlo) integration.

This acquisition function has several properties that make it especially useful for optimizing with online experiments---it robustly handles noise and constraints, is easily extended to batch optimization, and is easy to implement and maintain in production.  There are other acquisition functions that have been used for multi-fidelity problems, which we describe in Section \ref{sec:discussion}.

We adapt noisy EI to the multi-task setting by restricting the definition of the best-feasible policy to include only observations from the primary task (online observations). Given online observations at $X_T$ and simulator observations at $X_S$, we now let $\mathcal{D}_T$ represent online observations and $\mathcal{D}_S$ simulator observations. We then use as the acquisition function
\begin{equation}\label{eq:acq}
    \alpha_{\textrm{NEI}}(\mathbf{x} | X_T, \mathcal{D}_T \cup \mathcal{D}_S).
\end{equation}
Simulator observations are thus used to improve the model posteriors $f(X_T)$ and $\mathbf{c}(X_T)$, but are not considered to have been an observation for the purposes of evaluating improvement.

\subsection{The Bayesian Optimization Loop}\label{sec:bo_loop}
Optimizing the acquisition function produces a batch of points that maximizes the expected improvement over the policies that have been observed online, in expectation over the model predictions for their online outcomes. We then have the choice to launch the batch either online or to the simulator. Here we used a fixed strategy where batches of online experiments and batches of simulator experiments are interleaved. The full Bayesian optimization loop used here is given in Algorithm \ref{algo:bo}.

\begin{algorithm}[htb]\label{algo:bo}
\KwData{Online batch size $n_T$, simulator batch size $n_S$, optimization batch size $n_o \geq n_T$, number of optimization iterations $K$.}

Construct a quasi-random batch of $n_T$ policies to be tested online, $X_T$, and $n_S$ policies to be tested on the simulator, $X_S$.

Run the online and simulator tests. Let $\mathcal{D}$ represent the complete collection of observations.

\For{$k = 1, \ldots, K$}{
Fit an MTGP to the observations $\mathcal{D}$ for each outcome of interest.

Optimize the acquisition function (\ref{eq:acq}) to obtain a batch of $n_o$ optimized policies to be tested.

Test the $n_o$ optimized policies on the simulator. Let $\mathcal{D}_{S, k}$ represent the observations and update $\mathcal{D} = \mathcal{D} \cup \mathcal{D}_{S, k}$.

Re-fit the MTGPs for each outcome to incorporate the new simulator observations.

Select the best $n_T$ of the optimized policies to be tested online. Let $\mathcal{D}_{T, k}$ represent the observations, and update $\mathcal{D} = \mathcal{D} \cup \mathcal{D}_{T, k}$.
}
\Return{The best policy observed online.}
\caption{Online-Offline Bayesian Optimization}
\end{algorithm}

In our experiments here, the initial quasi-random batches consisted of $n_T=20$ online tests and $n_S=100$ simulator tests. The initial batches do not necessarily have to cover the same space---the simulator can be used to explore policies on a wider design space than the online tests, allowing for more exploration while not veering too far off-policy in the online tests. In Line 4, for the first iteration ($k=1$), we fit a 2-task MTGP as done in Section \ref{sec:mtgp_emp}. For later iterations where there were multiple batches of simulator observations, we introduced additional tasks to handle non-stationarity across simulator batches, which are handled with a low-rank ICM kernel. This aspect of the modeling is described in Appendix \ref{sec:nonstationary}. In Line 5, we used the acquisition function to generate a batch of $n_o=30$ optimized policies. In Line 8, we used Thompson sampling \citep{thompson33} to select 8--10 policies for online tests, depending on available capacity.

Interleaving online and simulator batches as is done in Algorithm \ref{algo:bo} improves the quality of the online tests, and thus user experience in the test groups, by reducing the predictive uncertainty of candidates that have high EI. EI measures a balance between exploration and exploitation, so a policy selected in the first acquisition function optimization (Line 5) may have high EI because it has high predictive uncertainty. Running that policy on the simulator (Line 6) will likely reduce that uncertainty and can filter out from the online batch policies that are no longer likely to be good after their simulator observation. The online batch can thus be more exploitative (lower regret), while still maintaining thorough exploration through the simulator. As an alternative to interleaving the batches, one could use the acquisition function to weigh the relative efficiency and resource requirements of online and simulator experiments---we discuss this further in Section \ref{sec:discussion}.

In Online Appendix 2, we apply Algorithm \ref{algo:bo} to a synthetic online-offline problem that is modeled after our application here. There we compare to regular, single-task BO, and also evaluate how much improvement comes from the interleaved offline batches (Lines 6 and 7). Our focus here, however, is on real online optimizations, which we turn to now.

\subsection{Examples of Value Model Tuning with Online-Offline Experiments}

We now provide five examples of Bayesian optimization using both online and simulator observations according to the loop in Algorithm \ref{algo:bo}. Each of these experiments sought to maximize a particular objective outcome of the recommendation system's online performance, subject to 3--5 constraints on other outcomes. The constraint outcomes here are other metrics that have trade-offs with the objective, so improving one can often be to the detriment of the other. The experiments targeted different sets of value model parameters, with the design space dimensionality ranging from 10 to 20.

\begin{figure}[tb]
\centering
\includegraphics{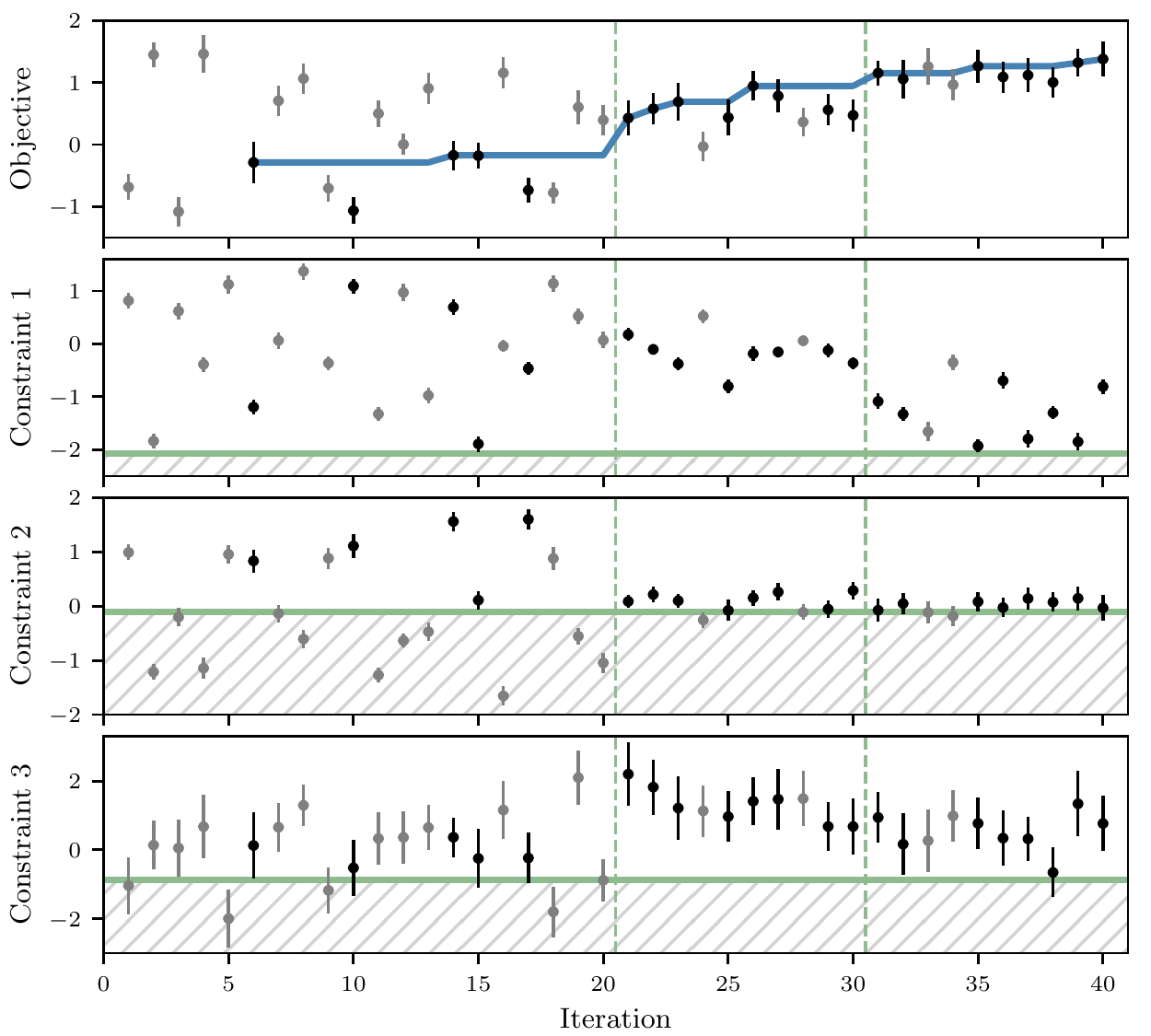}
\caption{Results of a multi-task optimization to maximize the objective subject to three constraints. Each point shows an online observation (mean and two standard errors). Gray points were infeasible in expectation, and the infeasible region for each constraint is shaded. The solid line for the objective tracks the expected best-feasible objective. Dashed vertical lines delineate batches of policies evaluated in parallel. A simulator batch was run prior to each online batch. Bayesian optimization with the MTGP was able to find significantly better policies with a limited number of online tests.}
\label{fig:optimization}
\end{figure}

\begin{figure}[tb]
\centering
\includegraphics{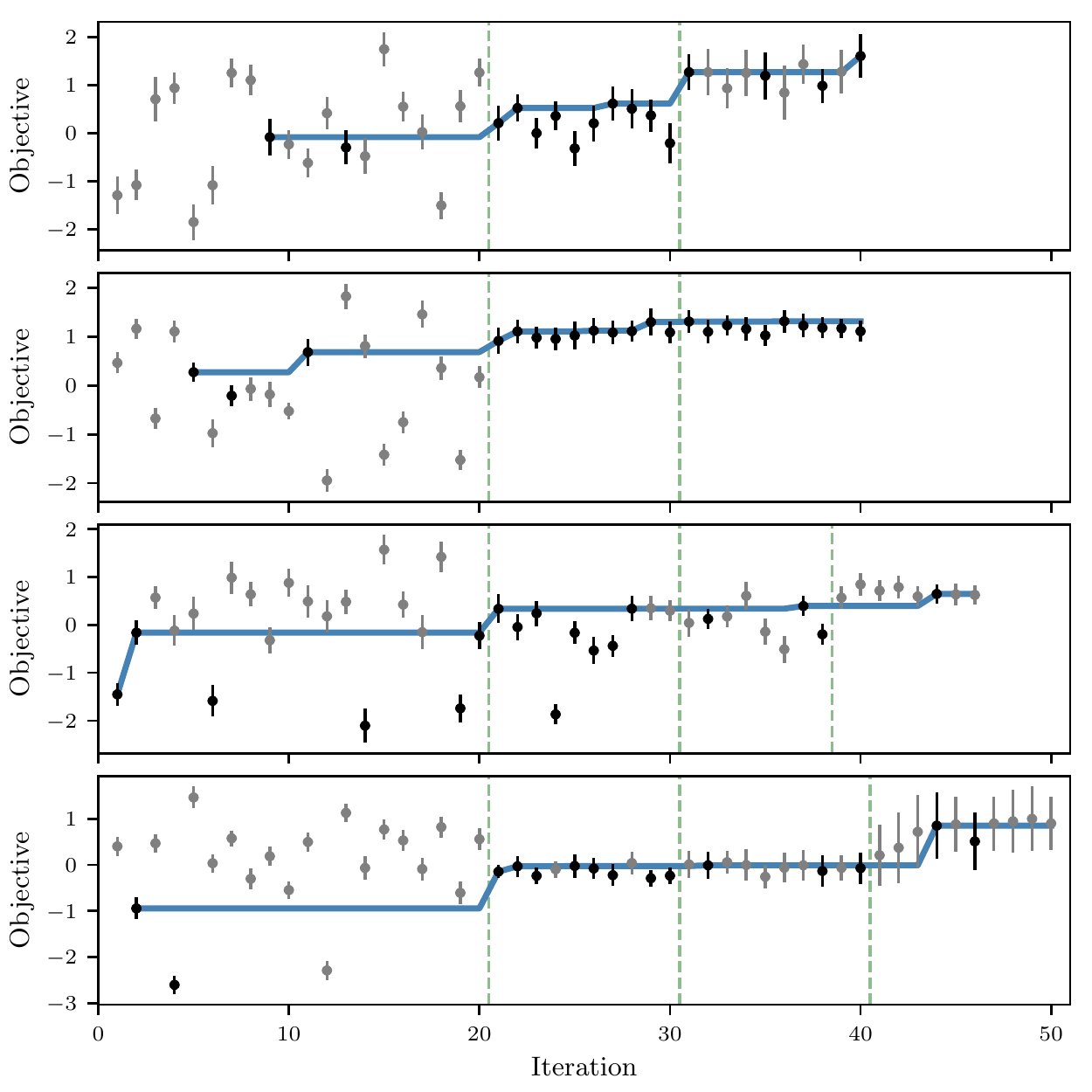}
\caption{Results of four separate policy searches over different sets of value model parameters. Each maximized the objective shown, subject to 3--5 constraints (not shown). As in Fig. \ref{fig:optimization}, grayed-out points were infeasible in expectation, the solid line tracks the expected best-feasible policy, and vertical lines delineate batches of policies evaluated online in parallel. These examples show the success of multi-task Bayesian optimization in 10--20 dimensions.}
\label{fig:four_optimizations}
\end{figure}

Fig. \ref{fig:optimization} shows the results of a round of Bayesian optimization in a 13-dimensional space with three constraints. The quasi-random initialization of 20 online observations and 100 simulator observations was followed by two rounds of interleaved simulator and online batches, as described in Section \ref{sec:bo_loop}, for a total of 160 policies evaluated in the simulator and 40 policies evaluated online. Outcomes and constraints were each modeled with an MTGP. The figure shows the results of each online observation. Most of the policies in the initialization were infeasible in expectation, and none of the feasible policies improved over the status quo value of 0. The status quo policy had previously been manually tuned and so quasi-random search was not sufficient to improve the objective. Optimizing with the MTGP, on the other hand, led to significantly better feasible policies in the optimized batches. Not only did the optimization improve the best-feasible objective, it also increased the proportion of online observations that satisfied the constraints, which corresponds to improved user experience in the test groups. The second optimized batch continued to improve over the first as it was able to push closer to the constraint bounds. The fact that we are able to jointly optimize 13 parameters with only 40 total online tests is remarkable, and is entirely due to the MTGP and the simulator.

The optimization gains in Fig. \ref{fig:optimization} are typical of a collection of successful optimizations we have done using the MTGP with the simulator. Fig. \ref{fig:four_optimizations} shows the results of four other optimizations, with 3--5 constraints and 10--20 parameters. As in Fig. \ref{fig:optimization}, for these problems the constraints were restrictive enough that most of the initial policies were infeasible. Subsequent optimized batches found feasible policies that substantially improved the objective value. For all of these optimizations the total number of online observations required was greatly reduced compared to single-task Bayesian optimization: 40--50 total online observations for 10--20 parameters.

\section{MTGP Generalization and the Value of a Simulator Observation}\label{sec:generalization}

We have seen that the MTGP is able to make accurate predictions of online outcomes using a small number of online observations and a large number of biased offline observations. The model is essentially able to use offline observations as a substitute for online observations. We now study the extent to which simulator observations can replace online tests, and in particular the number of simulator observations required and the conditions in which the model can effectively borrow strength from the simulator observations. Both of these are ultimately a question of MTGP generalization behavior, which we now describe first theoretically and then empirically.

\subsection{Learning curves for the MTGP}\label{sec:theory}

We now define the quantities used to evaluate MTGP generalization and discuss their theoretical properties. We begin with the single-task setting of (\ref{eq:gp_posterior}) where we have observed data $\{X, y\}$ and make a prediction at point $\mathbf{x}_*$. We suppose that there is some probability density from which training points $X$ and test points $\mathbf{x}_*$ are drawn, i.i.d. \citet{rasmussen06} show that if the true $f$ is a draw from a GP, then squared prediction error equals the predictive variance:
\begin{equation*}
    \mathbb{E}_{f} \left[ \left(\mu(\mathbf{x}_*, X, \mathbf{y}) - f(\mathbf{x}_*) \right)^2 \right] = \sigma^2(\mathbf{x}_*, X).
\end{equation*}
We define the single-task \textit{learning curve} for $n$ training points as the expected squared prediction error, which is
\begin{equation}\label{eq:st_learning}
    \epsilon(n) = \mathbb{E}_{X, \mathbf{x}_*} \left[ \sigma^2(\mathbf{x}_*, X) \right],
\end{equation}
for $X$ of size $n$.

These definitions extend naturally to the multi-task case, where each point in $X$ has an associated task and there is a target task associated with $\mathbf{x}_*$. For the purposes of the theoretical results, for the remainder of this section we will focus on the two-task case described by \citet{chai09}. There is a ``primary" task (online observations) for which we make predictions and a ``secondary" task (simulator observations) which can be used to improve the predictions on the primary task. For the theoretical results here, we assume that the two tasks have the same output variance, which is then pushed into the parameter kernel $\kappa(\mathbf{x}, \mathbf{x}')$ so that
\begin{equation*}
    B = \begin{pmatrix}
    1 & \rho \\
    \rho & 1
    \end{pmatrix}.
\end{equation*}
We do not make this assumption when actually using the model in the other sections of this paper, in which the task output variances are inferred from the data---it is only for the purposes of deriving the bound in Proposition \ref{prop:bound}.

We suppose that we have training observations at $X_T$ on the primary task and $X_S$ on the secondary task, with $n_T$ and $n_S$ the sizes of $X_T$ and $X_S$ respectively. The posterior predictive variance $\sigma_T^2(\mathbf{x}_*, \rho, X_T, X_S)$ can then be computed in closed form by substituting the multi-task kernel into (\ref{eq:kriging_b}).

The multi-task learning curve for $n_T$ primary task observations and $n_S$ secondary task observations is defined in the same way as the single-task learning curve,
\begin{equation}\label{eq:mt_learning}
    \epsilon_T(\rho, n_T, n_S) = \mathbb{E}_{X_T, X_S, \mathbf{x}_*} \left[ \sigma_T^2(\mathbf{x}_*, \rho, X_T, X_S) \right].
\end{equation}
\citet{chai09} proved an elegant bound on the multi-task predictive variance that provides insight into the relative value of online and offline observations for predictive accuracy.
\begin{proposition}{\citep[Proposition 5(a)]{chai09}}\label{prop:bound}
\begin{equation*}
    \sigma_T^2(\mathbf{x}_*, \rho, X_T, X_S) \geq \rho^2 \sigma_T^2(\mathbf{x}_*, \rho=1, X_T, X_S) + (1 - \rho^2) \sigma_T^2(\mathbf{x}_*, \rho=0, X_T, X_S).
\end{equation*}
\end{proposition}
When $\rho=1$, observations from each task are equivalent, and the model reduces to a single-task GP with the combined training set $X=[X_T, X_S]$. Similarly, with $\rho=0$ the secondary task observations are ignored, and the model reduces to a single-task GP but with $X=X_T$. Taking the expectation over i.i.d. $X_T$, $X_S$, and $\mathbf{x}_*$ yields the following result.
\begin{corollary}\label{cor:bound} Let $\epsilon_{\kappa}(n)$ be the learning curve for the single-task GP with covariance function $\kappa(\mathbf{x}, \mathbf{x}')$. Then,
\begin{equation*}
\epsilon_T(\rho, n_T, n_S) \geq \rho^2 \epsilon_{\kappa}(n_T + n_S) + (1 - \rho^2) \epsilon_{\kappa}(n_T).
\end{equation*}
\end{corollary}
The result in Corollary \ref{cor:bound} is a bound, however \citet{chai09} showed empirically that it is relatively tight and closely approximates the multi-task learning curve. This result shows that the ability of the multi-task model to generalize depends on the ability to generalize on each task separately, $\epsilon_{\kappa}(n)$,  and the inter-task correlation $\rho$. When tasks are highly correlated, the $\rho^2 \epsilon_{\kappa}(n_T + n_S)$ term dominates and secondary task observations propel the learning forward on the single-task learning curve. Alternatively, when task correlation is low the $(1 - \rho^2) \epsilon_{\kappa}(n_T)$ term dominates and secondary task observations have little effect.

Besides providing insight into the factors behind generalization, a useful property of the bound is that the quantities on the right-hand-side can be empirically estimated without primary task observations. The single-task learning curve $\epsilon_{\kappa}(n)$ can be estimated from secondary task observations only. Given an estimate of $\rho^2$ (perhaps from past usage of the simulator), we can then use the bound to understand the expected generalization to the primary task before even making primary task observations.

\subsection{Empirical Learning Curves}\label{sec:empirical_lc}

\begin{figure}
\centering
\includegraphics{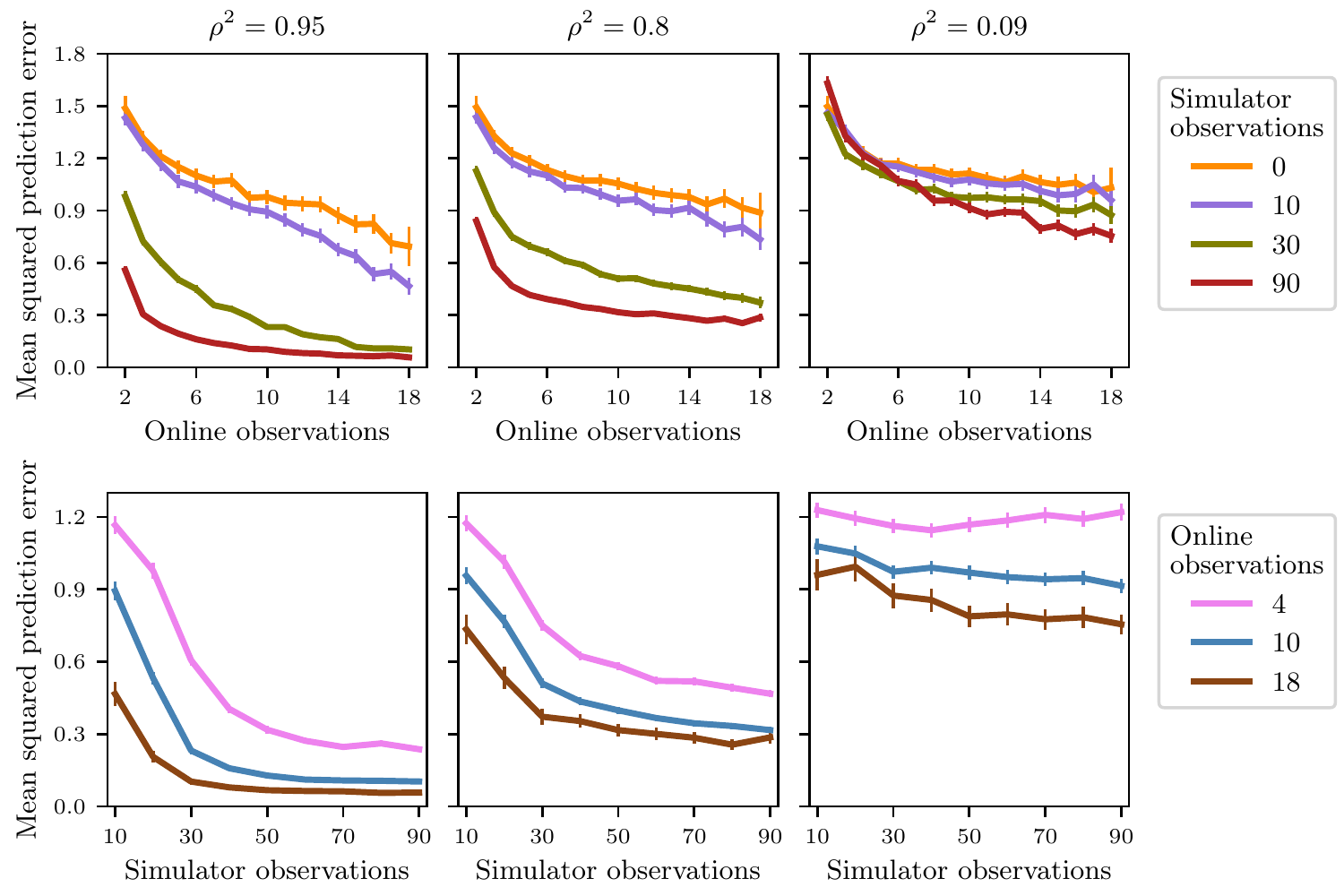}
\caption{Empirical learning curves from value model tuning experiments for the MTGP. Each column shows an outcome with different levels of squared inter-task correlation $\rho^2$ as indicated. Curves show the average over 500 training set draws, with two standard errors as error bars. Learning curves are shown as a function of the number of online observations $n_T$ with number of offline observations $n_S$ fixed (top row), and as a function of $n_S$ with $n_T$ fixed (bottom row). Simulator observations greatly accelerate learning when the inter-task correlation $\rho^2$ is high.}
\label{fig:learning_curves}
\end{figure}

We use cross-validation to estimate empirical learning curves $\epsilon_T(\rho, n_T, n_S)$ for the value model policy models in Section \ref{sec:mtgp_emp}. To estimate the MTGP learning curve at $n_T$ online observations and $n_S$ simulator observations, we randomly sampled (without replacement) $n_T$ of the 20 online observations and $n_S$ of the 100 simulator observations and fit a model to only those data. We then made predictions at the held-out $20-n_T$ online points and evaluated mean squared prediction error and mean predictive variance. This was repeated with 500 samples of the data to approximate the expectation over $X_T$, $X_S$, and $\mathbf{x}_*$ in (\ref{eq:mt_learning}). Empirical learning curves are shown in Fig. \ref{fig:learning_curves} for three outcomes from an experiment over a 10-d parameter space. These outcomes were chosen to illustrate learning curves with weak, medium, and strong inter-task correlations; the square of the inter-task correlation, $\rho^2$, is indicated for each outcome.

The empirical learning curves show how the simulator observations significantly improve generalization throughout the learning curve. As expected from theory, the extent to which simulator observations improve learning depends on the squared inter-task correlation $\rho^2$. For $\rho^2$ high, 2 online observations combined with 90 simulator observations produce a better model than 18 online observations. For $\rho^2$ small, at least 5 online observations are needed before the simulator even begins to provide benefit. The rate at which prediction error decreases with simulator observations is much lower than with high correlation. Comparing the $\rho^2=0.8$ and $\rho^2=0.09$ outcomes in Fig. \ref{fig:learning_curves} with $0$ simulator observations shows that the online response surfaces are of a similar difficulty to predict with the GP; the dramatic difference in the response to simulator observations comes from the difference in inter-task correlation.

\subsection{Comparing Theory to Empirical Learning Curves}\label{sec:emp_vs_theory}

\begin{figure}[tb]
\centering
\includegraphics{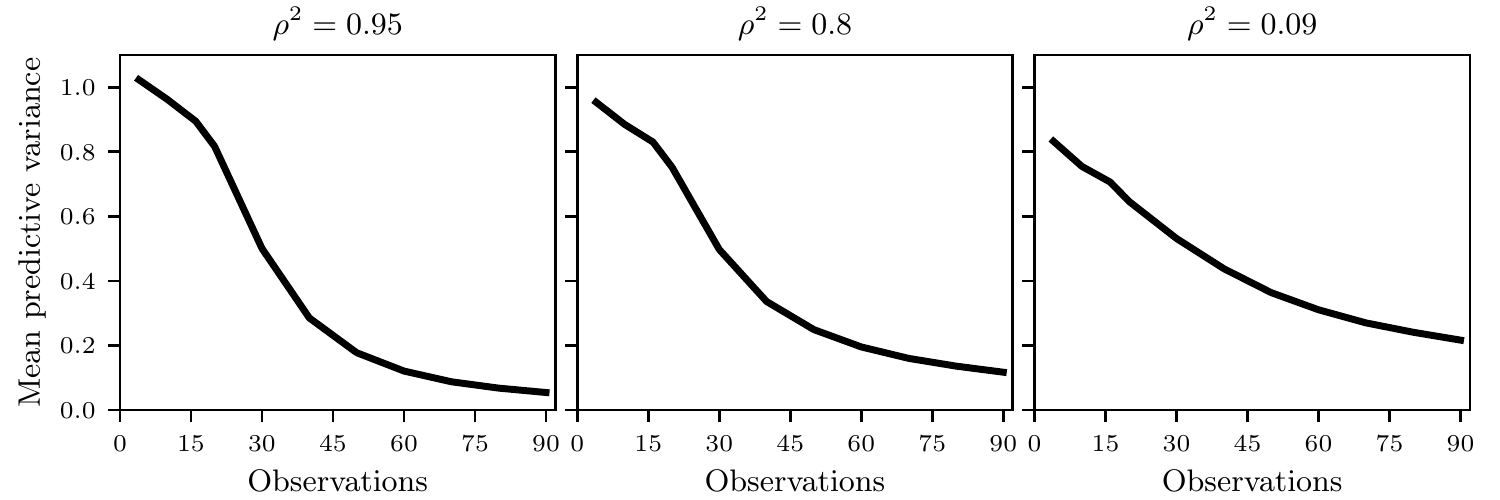}
\caption{Single-task empirical learning curves estimated from simulator observations only. Panels correspond to the same three outcomes as Fig. \ref{fig:learning_curves}, in a 10-d parameter space. There is an initial slow phase of learning in which there are not enough observations to infer the kernel, after which learning accelerates.}
\label{fig:single_task_learning}
\end{figure}

The theoretical result in Corollary \ref{cor:bound} provides an elegant relationship between the single-task learning curve, $\epsilon_{\kappa}(n)$, and the multi-task learning curve. Under the ICM assumption of a shared covariance function, the single-task learning curve can be estimated using only simulator observations. To estimate the learning curve, we use the same cross-validation procedure used for the MTGP learning curves in Fig. \ref{fig:learning_curves}: $n$ simulator observations were chosen at random, a single-task GP was fit to those data, and predictive variance was measured on the $100-n$ held-out points. Fig. \ref{fig:single_task_learning} shows the single-task learning curves for the three outcomes from Fig. \ref{fig:learning_curves}. 

\begin{figure}[tb]
\centering
\includegraphics{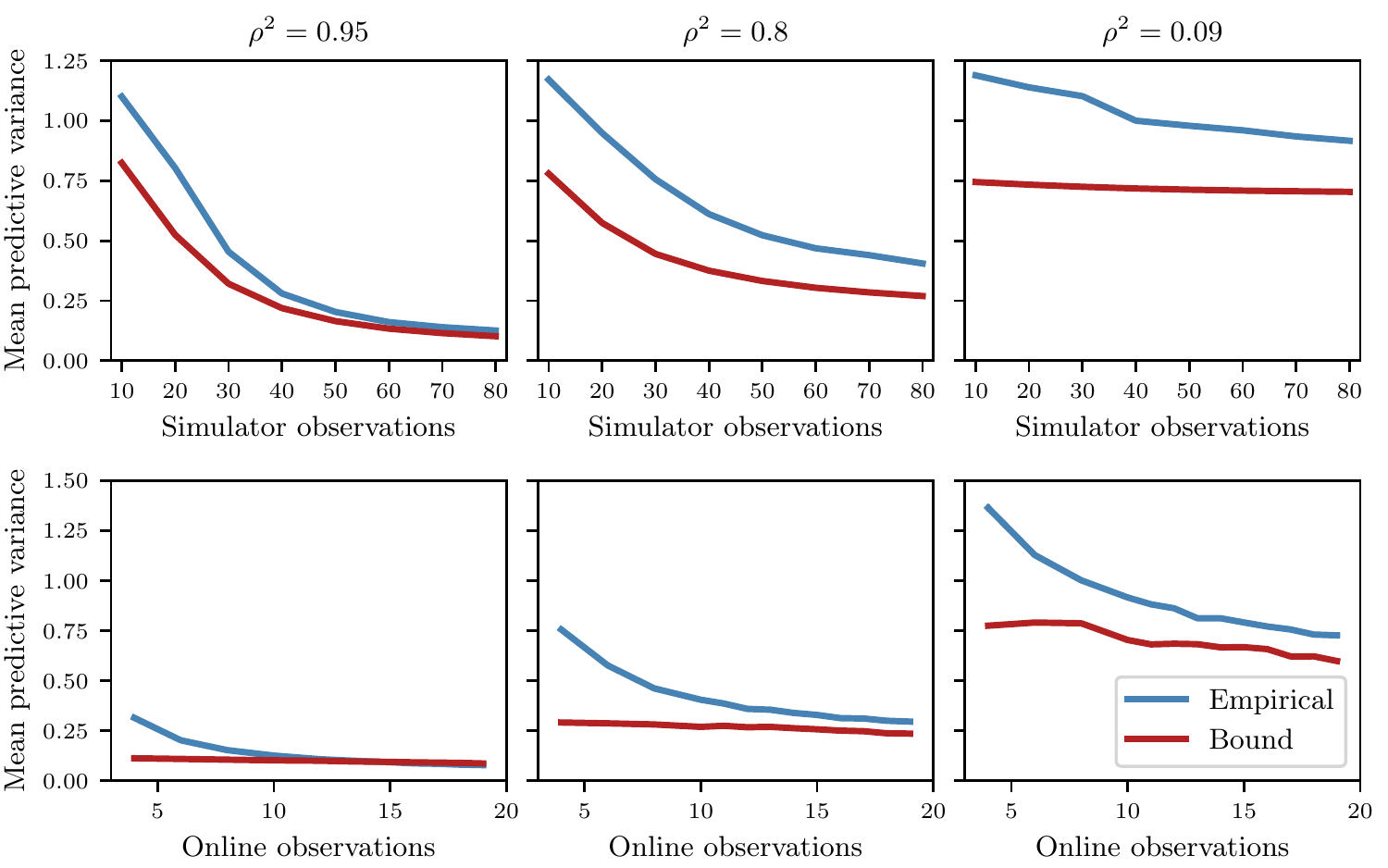}
\caption{Empirical multi-task learning curves compared to the theoretical bound applied to the empirical single-task learning curves. Top row shows learning curves as a function of the number of online observations $n_T$ with the number of simulator observations $n_S$ fixed at 80. Bottom row shows learning curves as a function of the number of simulator observations $n_S$, with $n_T=10$ online observations. Theoretical results provide a useful indication of real-world performance.}
\label{fig:emp_bound}
\end{figure}

The single-task learning curves are characterized by an initial slow period of learning, followed by a rapid and saturating reduction in predictive variance. The initial slow period can be attributed to the difficulty in inferring the kernel in 10 dimensions until a minimum number of observations have been obtained. Together with the estimated $\rho^2$, these learning curves are used to evaluate the bound $\rho^2 \epsilon_{\kappa}(n_T + n_S) + (1 - \rho^2) \epsilon_{\kappa}(n_T)$. Importantly, this is computed using only simulator observations and $\rho^2$. This bound is compared to the actual empirical multi-task learning curve $\epsilon_{T}(\rho, n_T, n_S)$ in Fig. \ref{fig:emp_bound}.

For these real experiments, although the MTGP fits the data well it is not the actual data generating process as assumed by the bound, and so we do not expect the bound to necessarily be satisfied by empirical learning curves. For instance, the real data do not have the same output variance for each task as assumed by Proposition \ref{prop:bound}. Although the output variances will be similar because we standardize the observations from each task prior to fitting, we do infer a separate output variance for each task. Despite these differences between the theoretical framing and practical setting, we do find here that the bound provides useful guidance for understanding the role of online and simulator observations in learning.

The empirical learning curves are generally consistent with theory, and in particular the role of $\rho^2$ on learning is well predicted by theory. With fewer than 10 online observations, the empirical learning curve shows a steeper decline than the bound. The source of this discrepancy can be seen by comparing the single-task learning curve of Fig. \ref{fig:single_task_learning} to the multi-task learning curves of Fig. \ref{fig:learning_curves}. In the single-task learning curve, the variance reduces slowly with $n$ up until around 20 observations. With simulator observations, the multi-task model avoids this slow initial phase of learning and exhibits a steep learning curve from the start. This suggests that improved kernel inference plays an important role in the MTGP performance, a factor that we analyze in the next section. As $n_T$ grows, the value and slope of the empirical learning curves are consistent with the theoretical bound.

These results show that theory can provide practical guidance on the number of online and simulator observations required to reach a desired generalization error, in addition to quantifying the value of incremental observations. In the single-task learning curves the $\rho^2=0.8$ outcome had higher mean predictive variance at $n=19$ than the $\rho^2=0.09$ outcome, however the bound correctly predicts that with simulator observations the $\rho^2=0.09$ outcome has higher mean predictive variance at $n_T=19$. Computing the bound for higher values of $n_T$ could guide a decision when trading off predictive variance with the resource requirements of online observations.

\subsection{Kernel Inference and MTGP Generalization}

The theoretical results and empirical learning curves show that simulator observations increase the effective amount of data, with the relative value of simulator and online observations depending on $\rho^2$. More data improves the model predictions in two distinct ways. First, the GP is a linear smoother, and having more data nearby the prediction point reduces predictive uncertainty. Second, the additional data improves inference for the kernel hyperparameters: the kernel output variance, lengthscales, and the task covariance matrix. All of the learning curves computed so far included kernel inference as part of the learning. That is, the kernel hyperparameters were inferred from the subset of data, and so the learning curves incorporate both of these sources of learning.

The single-task learning curves in Fig. \ref{fig:single_task_learning} showed an initial slow phase of learning due to kernel inference. Due to the ICM assumption of a shared spatial kernel, the MTGP is able to use offline observations to infer the kernel and thus avoid this initial slow phase.

\begin{figure}[h]
\centering
\includegraphics{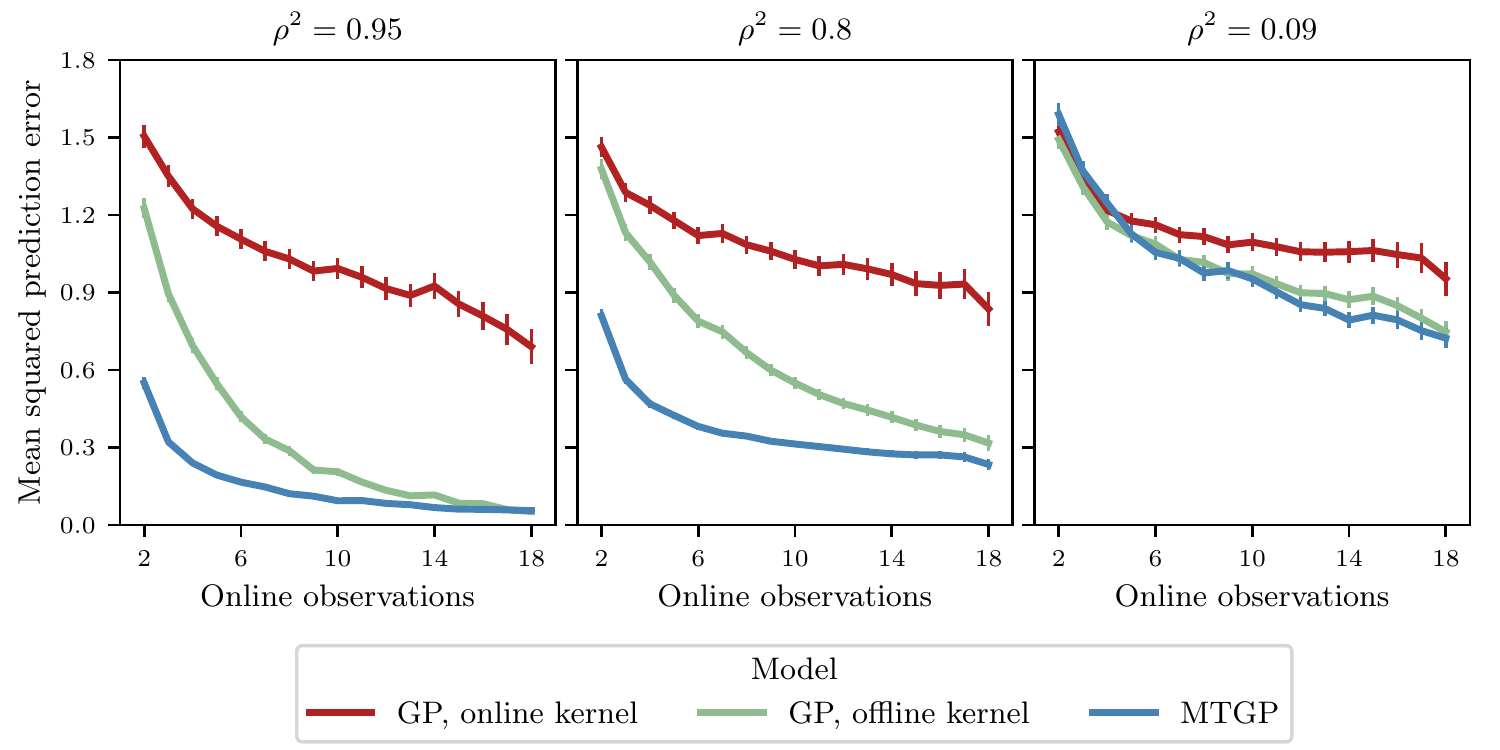}
\caption{Empirical learning curves for three models: A single-task GP fit to online observations, including kernel inference; a single-task GP that uses only online observations to compute the posterior but has the kernel fixed as that inferred from the simulator observations; and an MTGP with $100$ offline observations. For higher numbers of online observations, much of the performance improvement with the MTGP comes from the shared kernel with the offline task.}
\label{fig:kernel_lc}
\end{figure}

How much of the MTGP performance is due solely to improved kernel inference and the ICM assumption?  To answer this question, we computed empirical learning curves for three models, on the outcomes used in the previous section. The first model was a single-task GP fit only to the online data, with the kernel inferred as part of model fitting as usual. The third was an MTGP with the same online data plus $100$ offline observations, also with the kernel inferred as part of model fitting. The third model was a single-task GP that was given only online data for computing the posterior, however the kernel was fixed at the best offline kernel (that is, the kernel inferred when fitting a single-task GP to the $100$ offline observations). These empirical learning curves are shown in Fig. \ref{fig:kernel_lc}.

The two single-task GPs use the same data to compute the posterior, but we see that substituting the offline kernel for the kernel inferred from online data significantly accelerates learning. In fact, much of the gap in performance between the single-task GP and the MTGP can be attributed solely to improved ability to infer the kernel---not because of conditioning on more data in the GP posterior. For the outcomes in Fig. \ref{fig:kernel_lc} with medium and high values of $\rho^2$ there was still substantial value to the MTGP beyond just the offline kernel, but that diminished as the number of online observations grew. For the outcome with low $\rho^2$, practically all of the MTGP's performance improvement came from the kernel inference.

\section{Discussion}\label{sec:discussion}

We have provided empirical results and described practical techniques for using Bayesian optimization to tune an online machine learning system with the aid of a na\"{\i}ve simulator. Across a large set of experiment results, the MTGP with simulator observations provided remarkable gains in predictive performance over a single-task GP fit to online observations only. Section \ref{sec:mtgp_emp} showed that the single-task model failed badly to predict for most of the experiment outcomes we studied here---the optimizations in Section \ref{sec:optimization} would not have been possible without the simulator. These results showed that with the modeling power of the MTGP, even a simple, biased simulator provides a powerful way of enabling policy search via Bayesian optimization in settings where the limited pool for online experiments would otherwise be prohibitive.

Multi-fidelity Bayesian optimization has been studied with both continuous \citep{kandasamy17} and discrete \citep{kandasamy16} fidelities---here we had two fidelities, the online outputs and the simulator outputs. MTGPs have been used to model discrete fidelities in many physics and engineering applications \citep{wackernagel94, forrester2007, fernandez2016review}. \citet{Huang2006} first proposed Bayesian optimization with multiple fidelities using a heuristic extension to EI, albeit with a different kernel structure than the ICM kernel used here. In machine learning, the most common application of multi-fidelity Bayesian optimization is hyperparameter optimization with downsampled datasets \citep{swersky13}, an approach that has been developed with several kernels and acquisition functions \citep{klein15, klein17, poloczek17}. MTGPs have also been used for transfer learning in hyperparameter optimization, alongside a number of other models for learning relationships across tasks \citep{yogatama14, poloczek16, shilton17}. Other substantial machine learning applications of MTGPs and Bayesian optimization have been contextual bandit optimization \citep{krause11} and reinforcement learning \citep{chai08, metzen16, marco17}.

The empirical learning curves characterized the generalization of the multi-task model in practice, and showed the relative values of online and simulator observations. The theory behind MTGP identifies $\rho^2$ as the key factor determining the relative value of simulator observations. We found that our empirical learning curves were consistent with the theoretical bound; that $\rho^2$ was an important factor for multi-task learning, and that the theory could be used to guide decisions in sizing experiments. Previous empirical validation of the MTGP theory used simulations from the generating process. To our knowledge, these are the first results to compare empirics to theory on real-world multi-task modeling problems.

The ICM kernel was shown to be particularly effective at modeling the relationship between the online and offline observations, and enabling the simulator to improve predictions for online outcomes. The assumption of a shared spatial kernel allows kernel inference to rely primarily on the numerous simulator observations, and the implicit assumption of smooth bias was empirically appropriate for these experiments. Our results showed that a substantial part of the gain from using the MTGP can be attributed directly to the ICM shared spatial kernel. These results are consistent with work on Bayesian optimization for learning robot locomotion policies, where simulators were used for learning a custom kernel that was then transferred to real-world contexts for safe and sample-efficient learning \citep{antonova2017deep,rai2018using}.

Finally, we found that the MTGP allowed for effective optimization of a live recommendation system. We used noisy EI as the acquisition function because it naturally handles the properties of online experiments---noisy observations, noisy constraints, and large batches. EI does not, however, directly provide a way of selecting the modality of experimentation (online or simulator), hence we used a fixed strategy of interleaving simulator and online batches. Alternative acquisition functions could potentially identify whether an online batch or simulator batch should be done. Several heuristic extensions to EI allow for jointly selecting the point to be tested and the fidelity \citep{Huang2006, lam2015}. The knowledge gradient provides a principled way to measure the expected improvement in the online task that a simulator batch would provide \citep{poloczek17}. This could be compared to the improvement provided by an online batch, and the relative gains and efficiencies could be used to evaluate if it is most beneficial to run the simulator batch or the online batch. However, knowledge gradient has not been developed to handle constraints, a necessary requirement for our experiments. Predictive entropy search \citep{hernandez14, hernandez15, mcleod18} allows for measuring the effect a simulator batch has in reducing the entropy of the location of the online optimizer. However, it is not immediately clear how to construct a decision rule for evaluating the trade-off between information gain and experiment time. Building such a decision rule into the acquisition function remains an important area of future work.


\acks{We thank Si Chen, Yang Liu, Ashwin Murthy, and Pengfei Wang who worked on the simulator and systems integration. We also thank Brian Karrer for providing valuable modeling insights and David Arbour for work on model validation.  Finally, we thank Roberto Calandra, Mohammad Ghavamzadeh, Ron Kohavi, Alex Deng, and our anonymous reviewers for helpful feedback on this work.}

\appendix
\section{Handling Simulator Non-stationarity during Optimization}\label{sec:nonstationary}

As the optimization progresses, new batches of policies are constructed and launched either online or offline as we search for the optimal policy. In later iterations when we have accumulated several batches of simulator observations, optimization with the aid of the simulator has an additional challenge that must be handled in the modeling: non-stationarity across simulator batches.

For practical reasons, the sessions replayed on the simulator often cannot be a fully representative sample of sessions, rather they are sampled from live sessions during the simulator run.  This is similar to the ``live benchmarking" paradigm in \citet{bakshy15}, where requests are streamed to multiple servers in parallel.  Such live setups are often more efficient from a resource utilization perspective, since relevant state information can remain ``hot'' in distributed caches, and need not be saved to disk. This leads to a biased sample that is more likely to include sessions that occur at the same time as the simulator experiment. If the effects of the experiment are heterogeneous across people logged in at different times of the day or on different days, then we can expect the simulator outputs to have a bias that depends on when the simulation was run. This bias could be reduced through the use of propensity score weighting or regression adjustment \citep{sekhon11, bottou2013counterfactual,hartman2015sample}, but such approaches involve substantially more infrastructure. Moreover, the user population is not the only source of non-stationarity: there is also non-stationarity in the characteristics of items being scored by the machine learning system.

\begin{figure}[tb]
\centering
\includegraphics{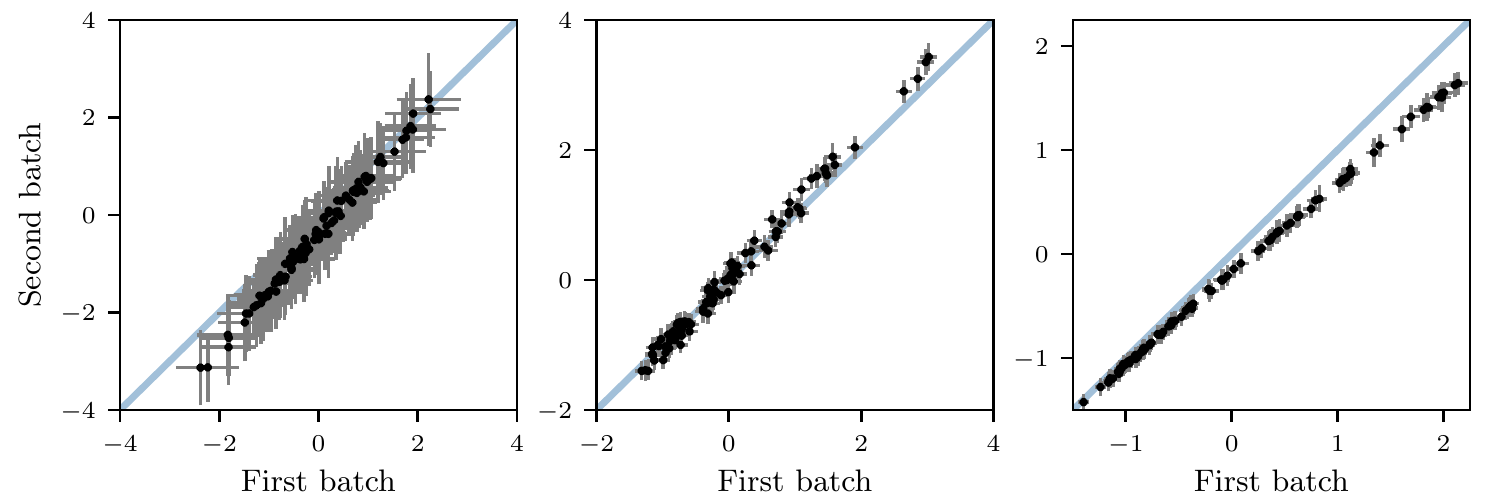}
\caption{Examples of the non-stationarity bias across batches run on the simulator, for three experiments. Each point was run on the simulator twice, in batches separated by several days. Sessions non-stationarity leads to differences in simulator outcomes when points are tested at different times.}
\label{fig:offline_shift}
\end{figure}

In online field experiments, these sources of non-stationarity are avoided by running the experiment for several days, as a matter of routine practice \citep{kohavi2007practical}. Offline experiments on the simulator are run for much less time to improve throughput, and so the bias across simulator batches must be corrected in the modeling. Fig. \ref{fig:offline_shift} shows an example of the effect of non-stationarity on simulator observations. This figure compares simulator observations from two one-hour periods of replayed sessions, separated by several days. Each experiment tested a batch of the same 100 points, and there are clear discrepancies in the simulator outcomes in observations at different times.

Bias across simulator batches can be handled naturally in the MTGP the same way the online-offline bias is handled, by treating each offline batch as a separate task. The need to treat each simulator batch as a separate task could make it challenging to estimate the $\mathcal{O}(D^2)$ cross-task covariances, however we can use a low-rank task covariance matrix as described in Section \ref{sec:mtgp}. We use a low-rank ICM kernel to learn adjustments for simulator batches, and then apply a full rank (rank-2) kernel to the online-offline tasks as done in Section \ref{sec:mtgp_emp}. 

Empirically, we find that the bias across simulator batches due to non-stationarity nearly always admits a low-rank representation. For example, the biases shown in Fig. \ref{fig:offline_shift} are all linear shifts of the response surface. Linear corrections are often used as multi-fidelity models in engineering \citep{fernandez2016review}, and correspond to a rank-1 ICM kernel, as discussed in \ref{sec:mtgp}. \citet{bonilla07} show that low-rank approximations such as that can be beneficial for generalization, and we use cross-validation to ensure that the model is performing well. In the Bayesian optimization loop in Algorithm 1, a subset of the initial points were repeated in every simulator batch so that there were repeated policies across batches to aid in learning this shift.

\vskip 0.2in
\bibliography{mtgp}

\newpage
\ShortHeadings{}{}
\begin{center}
\textbf{\large Supplemental Materials: Bayesian Optimization for Policy Search via  Online-Offline Experimentation}
\end{center}
\setcounter{section}{0}
\setcounter{equation}{0}
\setcounter{figure}{0}
\setcounter{page}{1}
\renewcommand{\theequation}{S\arabic{equation}}
\renewcommand{\thefigure}{S\arabic{figure}}
\renewcommand{\bibnumfmt}[1]{[S#1]}
\renewcommand{\citenumfont}[1]{S#1}

\section*{Online Appendix 1: Factors behind MTGP performance}

\begin{figure}[bh]
\centering
\includegraphics{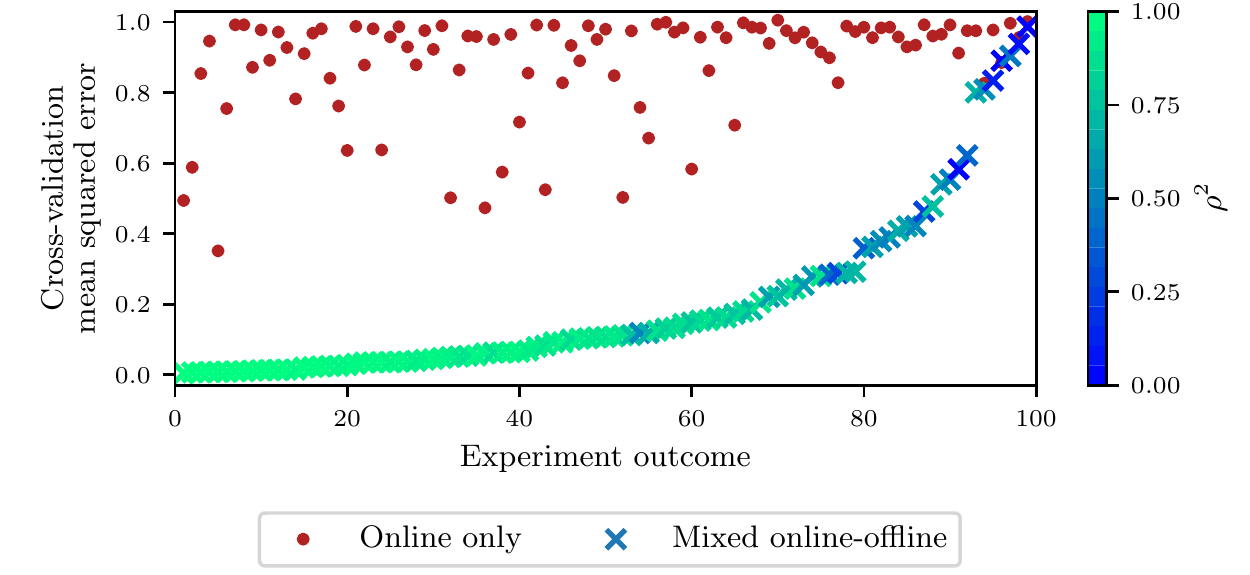}
\caption{The same results as Fig. \ref{fig:loo_meta} from the main text, with the MTGP markers colored according to the squared inter-task correlation, $\rho^2$. Outcomes with high MTGP error have moderate or low values of $\rho^2$. Low inter-task correlation is the factor behind high MTGP error.}
\label{fig:meta_rho}
\end{figure}

Section \ref{sec:mtgp_emp} of the main text evaluated cross-validation mean squared error for a single-task GP and an MTGP on 99 separate experiment outcomes. Fig. \ref{fig:loo_meta} showed that for many experiment outcomes, use of the simulator and the MTGP significantly improved prediction ability over the single-task GP. Section \ref{sec:generalization} showed that the key quantity in MTGP generalization is the squared inter-task correlation, $\rho^2$. Consistent with that, Fig. \ref{fig:meta_rho} shows that most outcomes had high values of $\rho^2$, but the outcomes with high MTGP MSE nearly all have low values of $\rho^2$.

\begin{figure}[tb]
\centering
\includegraphics{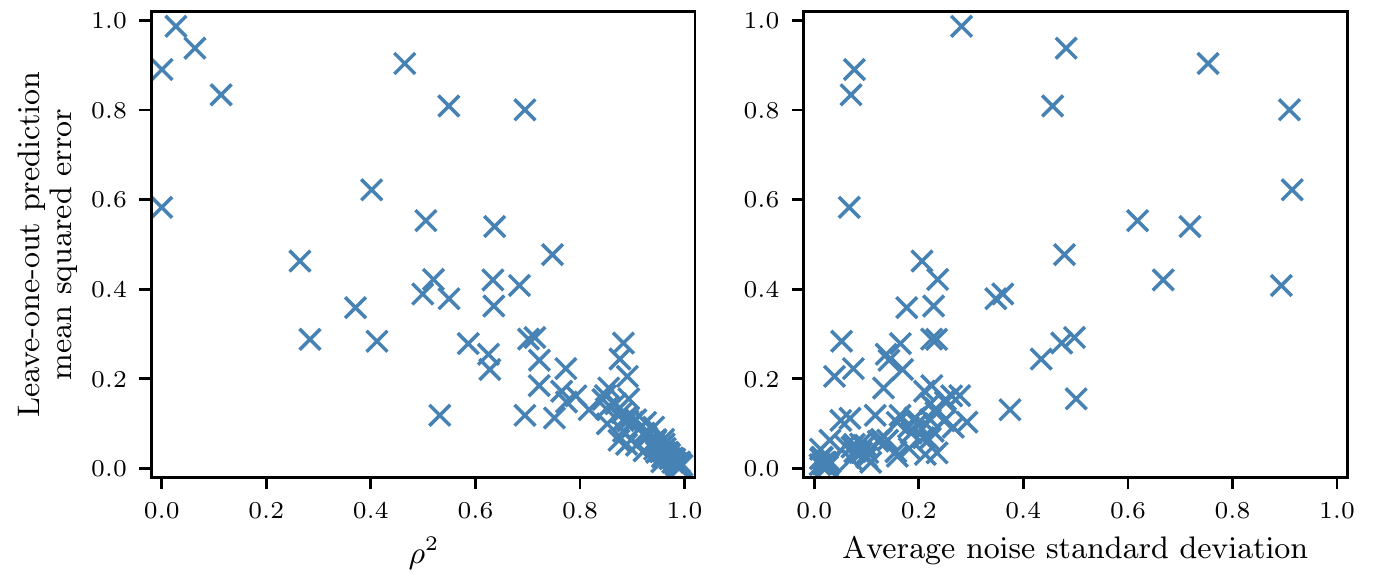}
\caption{(Left) $\rho^2$ vs. cross-validation error for the 99 experiment outcomes in Fig. \ref{fig:meta_rho}. As expected from theory, MTGP performance depends strongly on $\rho^2$. (Right) Cross-validation error vs. noise level for the same outcomes. Error tends to be higher for cases where the noise standard deviation is high, although there are some instances of high error with low noise level.}
\label{fig:rho_vs_noise}
\end{figure}

Fig. \ref{fig:rho_vs_noise} directly shows the relationship between $\rho^2$ and cross-validation MSE for those same 99 experiment outcomes in Fig. \ref{fig:meta_rho}. There is a clear correlation between the two, and values of $\rho^2$ greater than 0.75 always produced low MTGP MSE.

There are two potential sources of low inter-task correlation. The first is a high level of observation noise: If the level of observation noise is high relative to the effect size, the correlation will be low regardless of how similar the true response surfaces are. The second source of low inter-task correlation is poor fidelity of the simulator. To help identify the source of the low correlations, Fig. \ref{fig:rho_vs_noise} also compares cross validation error to the noise standard deviation. These results tell a mixed story: Many of the outcomes with low values of $\rho^2$ have high noise levels, suggesting noise as a likely cause, but others do not, suggesting poor simulator fidelity as the likely cause.

\section*{Online Appendix 2: MTBO performance on a synthetic problem}

The results in the main text focused on an empirical analysis of MTBO performance on real value model tuning experiments. Here we also provide a synthetic problem with characteristics similar to that of the real problems discussed in the text. The code to produce the results in this section is available at \url{http://ax.dev}.

The simulator contributes to the optimization in Algorithm 1 in two ways. The first is that after the random initialization of both online and offline observations (Line 2), we have seen that the MTGP has much lower prediction error than can be obtained without the simulator. Lower prediction error in the first optimized batch means that the acquisition function will select better points and reach the optimum more quickly. After the initialization, the simulator is used by interleaving offline and online batches. Optimized points selected by the acquisition function are tested first on the simulator, and then a filtered set based on those results are actually deployed online. This helps to increase the quality of the optimized batch prior to being observed online, which further accelerates the optimization.

The main goal of these tests on the synthetic problem are to gain insight into the relative contributions of these two factors. We thus compared three models: a single-task GP using only online points; an MTGP using Algorithm 1 as described in the main text, which used the offline task as a look-ahead by evaluating a larger set of candidates offline, updating the model, and launching online the subset that maximized utility; and finally an MTGP that used offline data for the initialization but did not interleave offline and online batches after that. That is, in the third model lines 6 and 7 of Algorithm 1 were skipped, and the optimization proceeded exactly as with the single-task GP, except for the offline observations in the initialization.

For the true, online surface we use the Hartmann 6 problem, a classic 6-parameter optimization test problem over the domain $x \in [0,1]^6$. We also simulated an additional outcome $g(x) = ||x||_2$ to be used as a constraint. As in the real application described in the main text, observations were made in batch either ``online" or ``offline." Online observations were direct function evaluations of the Hartmann 6 function $f(x)$ and constraint $g(x)$, with normally distributed noise added to each (standard deviation $0.1$). Offline observations were generated by applying a nonlinear transform to the Hartmann 6 surface:
\begin{equation*}
\tilde{f}(x) = \begin{cases}
\alpha_1 (x - m) + m, \quad \textrm{if } x \leq m,\\
\alpha_2 (x - m) + m, \quad \textrm{otherwise},
\end{cases}
\end{equation*}
where $m=0.75$, $\alpha_1 = 0.4$, and $\alpha_2 = 0.8$. Offline observations for the constraint, $\tilde{g}(x)$ were generated via the same transform with $m=1.25$, $\alpha_1 = 0.8$, and $\alpha_2 = 4$. Offline observations also had normally distributed noise added to them, with the same standard deviation $0.1$.

\begin{figure}[tb]
\centering
\includegraphics{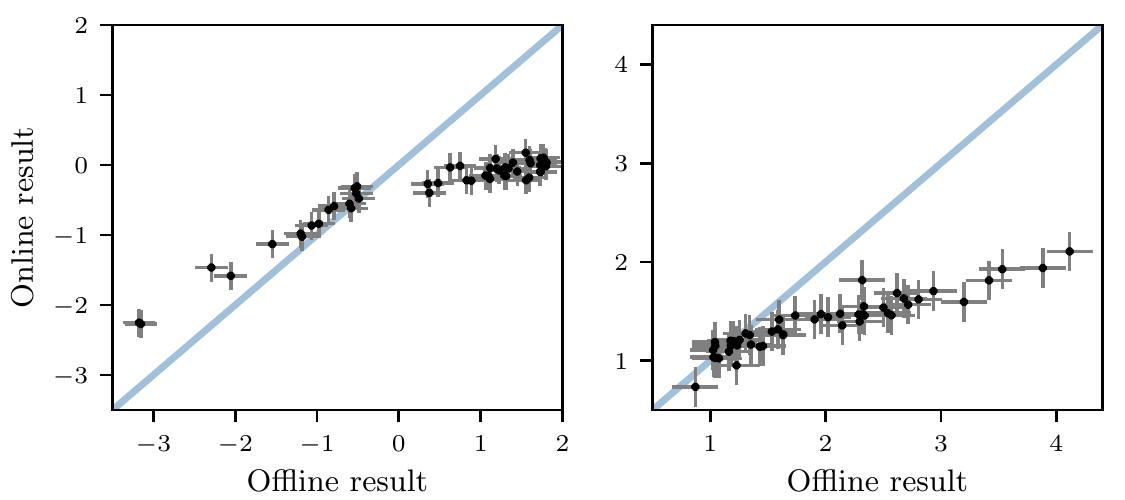}
\caption{A comparison between ``offline" and ``online" results for the synthetic problem objective (left) and constraint (right). The offline evaluations produce a biased estimate of the true, online surfaces. Observations from both tasks have noise.}
\label{fig:synthetic_bias}
\end{figure}

Fig. \ref{fig:synthetic_bias} shows a comparison between online and offline observations for this synthetic problem evaluated on a collection of points in the design space. The bias constructed in this problem is modeled after that seen in the real application in Fig. \ref{fig:online_offline} in the main text.

Bayesian optimization to minimize $f(x)$ subject to $g(x) \leq 5/4$ was done with the three models described above. All methods began with the same $n_T = 5$ online observations at points from a Sobol sequence. The MTGPs were additionally given $n_S = 20$ offline points for their initialization. After that, all methods made online observations in batches of $5$. For the MTGP using the loop in Algorithm 1, these were selected from an optimized batch of $n_o = 20$ points that were first tested offline. The optimization was done for a total of 20 online observations (4 batches of 5 points each), and was repeated 30 times, each with independent observation noise.

\begin{figure}[tb]
\centering
\includegraphics{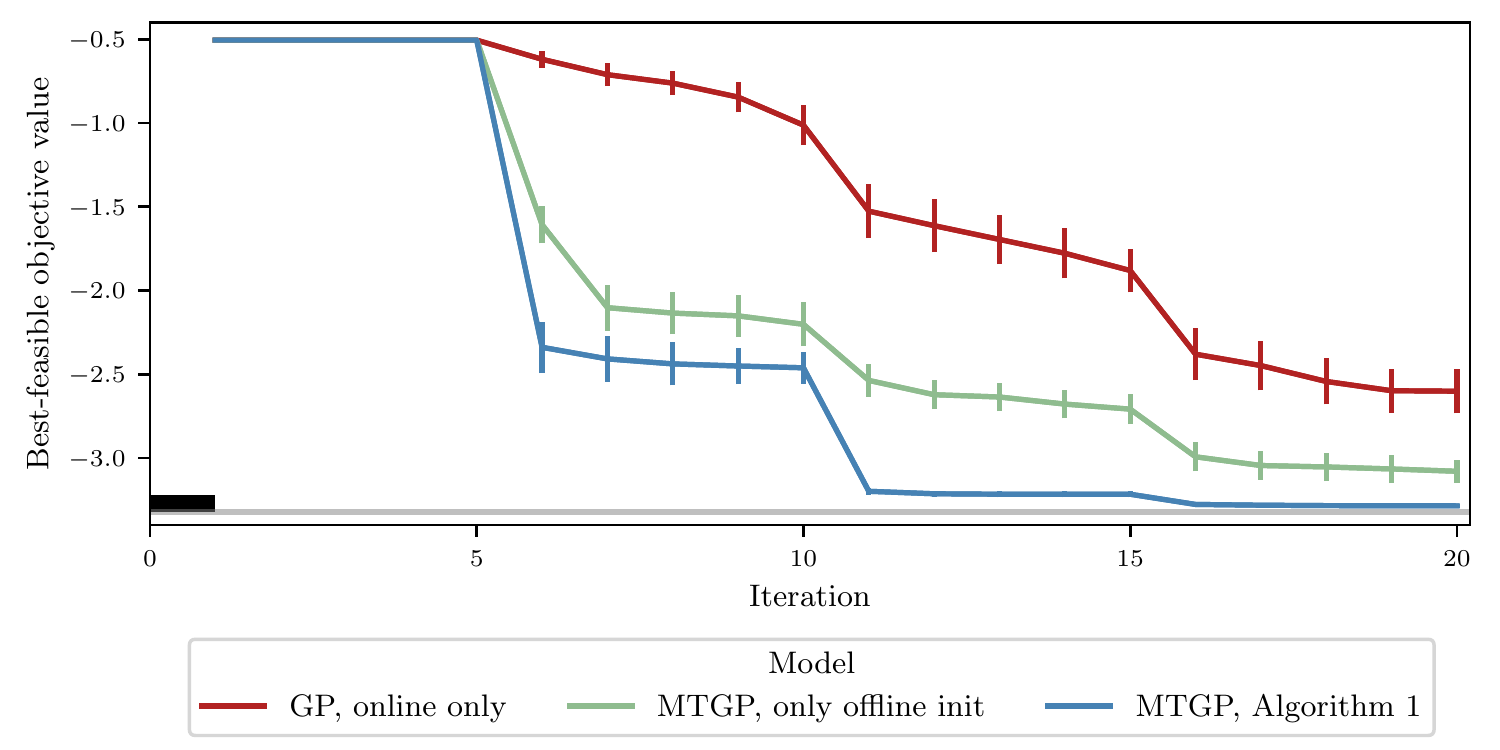}
\caption{Bayesian optimization performance (best feasible objective value by each iteration) on the synthetic minimization problem for three methods: A GP making only online observations, an MTGP that uses offline observations only for the initialization, and an MTGP using Algorithm 1 and interleaving offline and online observations throughout the optimization. Iterations are online observations, and values shown are the mean and two standard errors across 30 repeated optimizations. Horizontal gray line is the global optimum. Offline observations accelerate the optimization compared to optimization with only online observations, and there is significant value to interleaving offline observations throughout the optimization.}
\label{fig:synthetic_bo}
\end{figure}

Fig. \ref{fig:synthetic_bo} shows the results of the optimization for each model, as the mean (over 30 runs) best feasible point tested by each iteration (online observation). These results show that both uses of the simulator have significant value. Comparing the single-task GP to the MTGP with offline observations in the initialization, we see that the improved model predictions allow for better points to found much earlier in the optimization. Comparing the two MTGP models, we see that while the initialization plays an important role early on, using offline observations as a pre-test before online observations, as is done in Algorithm 1, leads to significantly better performance at all iterations.

\end{document}